%% file: combined.tex
\documentclass[11pt]{article}

\usepackage[utf8]{inputenc}
\usepackage[T1]{fontenc}
\usepackage{lmodern}
\usepackage{textcomp}
\usepackage{amsmath,amssymb}
\usepackage{graphicx}
\usepackage{longtable,booktabs,array}
\usepackage{calc}
\usepackage{caption}
\usepackage{xcolor}
\usepackage[margin=1in]{geometry}
\usepackage{placeins}
\usepackage{etoolbox}
\usepackage{float}                       
\usepackage{adjustbox}                   
\usepackage{fancyvrb}                    
\fvset{frame=single,framesep=4pt,fontsize=\small,xleftmargin=12pt,xrightmargin=12pt}
\usepackage{enumitem}                    
\setlist[itemize]{itemsep=2pt,parsep=2pt,topsep=4pt,leftmargin=*}
\setlist[enumerate]{itemsep=2pt,parsep=2pt,topsep=4pt,leftmargin=*}
\usepackage{multicol}                    
\usepackage{fancyhdr}                    

\newcommand{\citeref}[1]{\nolinebreak\,\textsuperscript{\textmd{#1}}}
\usepackage{hyperref}
\hypersetup{colorlinks=true,linkcolor=blue,urlcolor=blue,citecolor=blue,
            pdfborder={0 0 0}}
\usepackage{url}
\AtBeginEnvironment{longtable}{\footnotesize}
\AtBeginEnvironment{tabular}{\footnotesize}
\renewcommand{\arraystretch}{1.1}

\usepackage{setspace}
\AtBeginEnvironment{longtable}{\setstretch{1.0}}
\AtBeginEnvironment{tabular}{\setstretch{1.0}}

\DeclareRobustCommand{\_}{\textunderscore\allowbreak}

\usepackage[explicit]{titlesec}
\usepackage{needspace}                   

\titleformat{\section}{%
  \needspace{6\baselineskip}\Large\bfseries
}{}{0pt}{#1}
\titleformat{\subsection}{%
  \needspace{5\baselineskip}\large\bfseries
}{}{0pt}{#1}
\titleformat{\subsubsection}{%
  \needspace{4\baselineskip}\normalsize\bfseries\itshape
}{}{0pt}{#1}
\titleformat{\paragraph}[runin]{\normalsize\bfseries}{}{0pt}{#1}
\titlespacing*{\section}{0pt}{3.5ex plus 1.5ex minus .5ex}{1.6ex plus .4ex}
\titlespacing*{\subsection}{0pt}{2.4ex plus 1ex minus .4ex}{0.8ex plus .2ex}
\titlespacing*{\subsubsection}{0pt}{1.6ex plus 0.6ex minus .3ex}{0.5ex plus .1ex}
\titlespacing*{\paragraph}{0pt}{1.0ex}{1em}

\IfFileExists{microtype.sty}{\usepackage[protrusion=true,expansion=false]{microtype}}{}

\title{\bfseries\LARGE LifeSentence: Language models can encode human life course trajectories from longitudinal panel data.}
\author{Samuel Liu, Muchen Xi, William Yeoh, Joshua J. Jackson}
\date{May 5, 2026}

\makeatletter
\renewcommand{\maketitle}{%
  \begingroup
    \centering
    {\@title\par}\vspace{0.8em}
    {\large \@author\par}\vspace{0.4em}
    {\@date\par}\vspace{1.4em}
  \endgroup
}
\makeatother

\newcommand{\shorttitle}{LifeSentence: language models for life-course trajectories}
\fancypagestyle{plain}{%
  \fancyhf{}%
  \fancyhead[L]{\small\itshape\shorttitle}%
  \fancyhead[R]{\small\thepage}%
  \fancyfoot{}%
}

\begin{document}

\maketitle

\begin{center}
{\bfseries Abstract}
\end{center}
\vspace{-0.4em}
\begin{center}
\begin{minipage}{0.92\linewidth}
\begin{small}
\noindent Forecasting human life outcomes is important to gain insights into how individuals attain long and healthy lives. Conventional statistical approaches yield limited accuracy, potentially due to discarding the sequential structure of the life course. Modern methods such as transformer architectures require large scale training data that most longitudinal panel studies lack. Here we introduce LifeSentence, a model for life-course reasoning that bridges large language models with longitudinal panel data. By representing each life event as a structured natural-language record and instruction-tuning a pretrained 24-billion-parameter language model across an 18-task evaluation taxonomy spanning prediction, robustness and reasoning, LifeSentence supplements panel data with distributional knowledge already encoded during pretraining. Trained on approximately 65{,}000 individuals from the German Socio-Economic Panel---roughly 45 times fewer than prior transformer-based approaches---LifeSentence outperforms classical and deep learning baselines across all task families, achieving a threefold improvement in joint event-and-timing prediction from best baselines and 91.2\% Kendall's tau when reconstructing chronological order from timestamp-stripped event sets. Without explicit supervision, the model recovers documented patterns of social stratification, including the education premium, the gender wage gap and the motherhood penalty, from discrete event sequences alone. A natural-language interface further enables qualitatively new research queries, such as connecting an early-life history to a specified late-life endpoint, establishing LifeSentence as both a predictive tool and a probe for counterfactual exploration of human biographies.
\end{small}
\end{minipage}
\end{center}
\vspace{1.4em}

\section*{Introduction}

\input{article_body.tex}

\FloatBarrier
\clearpage

\renewcommand{\shorttitle}{LifeSentence \,---\, Supplementary Information}

\vspace*{2.5cm}
\begin{center}
  {\fontsize{28}{34}\selectfont\bfseries Supplementary Information\par}
  \vspace{1.0em}
  {\large\itshape Language models can encode human life course trajectories\\[2pt]from longitudinal panel data\par}
\end{center}
\vfill\null
\clearpage

\addcontentsline{toc}{section}{Supplementary Information}
\renewcommand{\thefigure}{S\arabic{figure}}
\renewcommand{\thetable}{S\arabic{table}}
\renewcommand{\theHfigure}{supp.\arabic{figure}}
\renewcommand{\theHtable}{supp.\arabic{table}}
\setcounter{figure}{0}
\setcounter{table}{0}

\input{supp_body.tex}

\end{document}

%% file: article_body.tex
Accurately forecasting the trajectory of a human life provides
foundational insights necessary to understand how individuals can
achieve long, healthy, and happy lives.\citeref{1--3} Yet
despite decades of inquiry, the predictive accuracy of standard models
remains modest,\citeref{4--6} suggesting that prevailing
analytical paradigms do not capture the high-dimensional, sequential
logic of human existence.
The necessity of studying life trajectories spans multiple
disciplines.\citeref{7,8} Shocks accumulate across the entire
lifespan, with critical periods linking early-life conditions to
late-life endpoints.\citeref{9,10} Cumulative advantage
specifies how small initial differences compound over
time,\citeref{11} while social clocks describe normative event
timing as a mechanism of social control.\citeref{12}
Collectively, these frameworks establish that a life is a dynamic
sequence shaped by timing, context, and accumulated
experiences.\citeref{13,14}

Traditional statistical tools discard sequential information, flattening
the rich temporal structure of the life course.\citeref{15,16}
The Fragile Families Challenge revealed that even sophisticated machine
learning struggled to predict life outcomes.\citeref{4}
Transformer architectures\citeref{17,18} offered a possible
solution.\citeref{19--21} The life2vec
study\citeref{22} validated this approach by training a
BERT-like transformer on registry data covering six million Danish
citizens, demonstrating that temporal event sequences carry predictive
signal about mortality and other life outcomes. However, life2vec
exposed a fundamental constraint: training transformers from scratch
requires large datasets.\citeref{23,24} Major longitudinal panel
studies track across a wider range of variables, but only across tens of
thousands rather than millions of participants.

Here, we introduce LifeSentence, a foundation model for life-course
reasoning that bridges large language models with longitudinal panel
data. Instead of learning from scratch, LifeSentence represents each
event as a structured natural-language description and instruction-tunes
a pretrained large language model on life-course prediction tasks.
Because the pretrained language model already encodes distributional
knowledge about the semantic relationships among concepts such as
education levels, occupations, and health conditions, fine-tuning only
needs to teach the model how they relate to one another across a
biography.\citeref{25--27}

LifeSentence outperforms traditional models across an 18-task evaluation
taxonomy\citeref{28} using approximately 65,000 individuals,
about 15 times fewer than life2vec. Beyond predictive accuracy,
LifeSentence reveals that the sequential structure of human lives
encodes far more information about social stratification than previous
analyses have extracted.\citeref{4,29} Without any explicit
supervision linking specific demographics to outcomes, the model's
generated trajectories reproduce well-documented patterns of
socioeconomic stratification, including the education income premium
previously established in the literature,\citeref{30,31} the
gender wage gap,\citeref{32,33} and the motherhood penalty from
event sequences alone,\citeref{34,35} indicating that the
co-occurrence structure of discrete life events is sufficient to recover
known patterns of socioeconomic stratification. These representations,
combined with a natural-language interface that enables qualitatively
new research queries ("generate all events until retirement," "connect
this early history to this late-life state"), establish LifeSentence as
both a predictive tool and a probe for what-if exploration of
life-course sequences.

\section*{Results}

\subsection*{Model Approach}

\begin{figure}[tbp]
{\centering
\includegraphics[width=\linewidth,keepaspectratio]{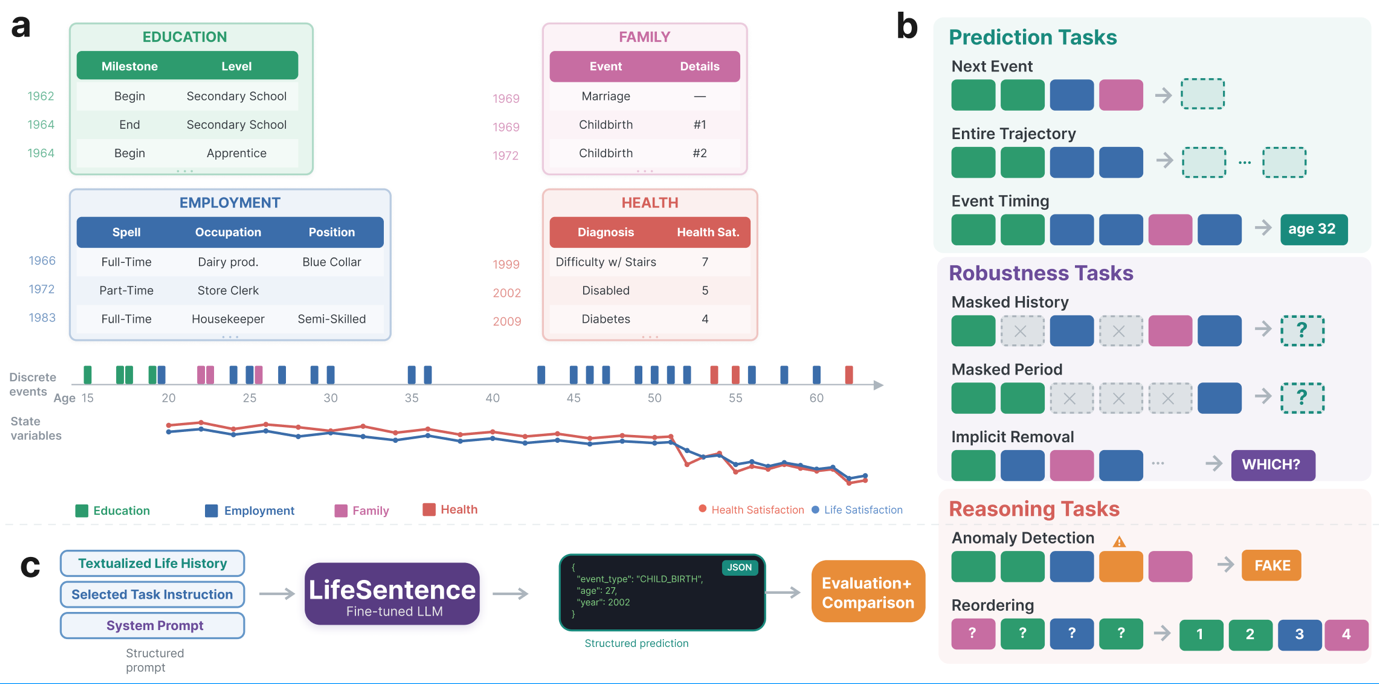}
\par}
\smallskip
{\small
\textbf{Figure 1\,---\, Overview of the LifeSentence framework.} (a)
A life sentence encodes an individual\textquotesingle s complete
biography as a temporally ordered sequence of discrete life events
spanning education, employment, family, and health domains. Each event
is recorded with its associated details, the
individual\textquotesingle s age, and the calendar year. Alongside
discrete events are periodic measurements of continuous state. (b) The
18-task evaluation taxonomy, organized into three families. Prediction
tasks assess forecasting capacity over varying horizons. Robustness
tasks test resilience to incomplete information through masked events,
redacted time periods, and silent deletions. Reasoning tasks probe
structural understanding of biography itself, including anomaly
detection and reordering. (c) The inference pipeline. A structured life
history, task-specific instruction, and system prompt are concatenated
into a single natural-language input. The fine-tuned model generates a
structured JSON prediction, which is evaluated against ground-truth
trajectories.
\par}
\end{figure}

We represent individual lives as temporally ordered sequences of
discrete life events---what we term LifeSentences---drawn from the
German Socio-Economic Panel (SOEP),\citeref{36,37} a
longitudinal household survey spanning decades of observation (Fig. 1a).
A LifeSentence encodes a biography: a woman completes secondary school
at 18, begins an apprenticeship, enters full-time employment as a nurse,
marries at 26, has two children, is diagnosed with hypertension at 52,
and retires at 63: each event recorded with its associated details, the
individual\textquotesingle s age, and calendar year. Events span
education (school transitions, certificates, degrees), employment
(occupational changes, position levels, retirement), family (marriages,
divorces, childbirths, widowhood), and health (diagnosed conditions,
death). Alongside discrete events, periodic measurements of continuous
state variables (income, life satisfaction, health satisfaction, and
occupational prestige) provide a complementary portrait of an
individual\textquotesingle s circumstances over time.

Each life event is represented not as a categorical index but as a
structured natural-language record, using occupation names, education
levels, and diagnostic terms that carry semantic content. This
representation allows the model to exploit distributional knowledge
already present in a pretrained large language
model\citeref{38,39}: the relationship between "apprenticeship"
and "vocational certificate" need not be learned from panel data alone.
We fine-tune Mistral Small 3.1 (24B parameters) using Low-Rank
Adaptation\citeref{40} on the 18 life-course prediction tasks
described below. A structured life history, task instruction, and system
prompt are concatenated into a single input; the model generates a
structured JSON prediction evaluated against ground-truth trajectories
(Fig. 1c).

LifeSentence is evaluated across an 18-task taxonomy organized into
three families of increasing inferential demand (Fig. 1b). Prediction
tasks\citeref{4,22} assess forecasting capacity: predicting next
events, generating full trajectories, forecasting to a given year,
estimating transition timing. Robustness tasks test resilience to
incomplete information by masking events, redacting time periods, or
silently removing observations: conditions mirroring gaps common in
panel data.\citeref{41,42} Reasoning tasks probe structural
understanding: detecting anomalous events, imputing silently removed
events, recovering chronological order from shuffled sequences without
timestamps, and translating discrete event sequences into the continuous
socioeconomic trajectories they produce.

\subsection*{LifeSentence predicts life events with high fidelity}

We first assessed LifeSentence by evaluating next-event prediction:
given an individual\textquotesingle s observed history truncated at a
random point, predict both the type and timing of the immediately
following event.

LifeSentence outperforms all baselines across every metric (Table 1).
The model achieves a joint accuracy of 35.3\% (correctly predicting both
the specific event type and the age at which it occurs), representing a
threefold improvement over the closest deep learning baseline and more
than double the next-best model. When the event type is predicted
correctly, LifeSentence places it within 1.15 years of the true age,
compared with 1.78 years or worse for all other baselines.

\begin{table}[tbp]
{\centering
\begin{adjustbox}{max width=\textwidth}
\begin{tabular}{@{}
  >{\raggedright\arraybackslash}p{(\linewidth - 6\tabcolsep) * \real{0.1528}}
  >{\raggedright\arraybackslash}p{(\linewidth - 6\tabcolsep) * \real{0.2180}}
  >{\raggedright\arraybackslash}p{(\linewidth - 6\tabcolsep) * \real{0.2180}}
  >{\raggedright\arraybackslash}p{(\linewidth - 6\tabcolsep) * \real{0.4112}}@{}}
\toprule
\textbf{Model} & \textbf{Event Accuracy} & \textbf{Joint Accuracy} &
\textbf{Conditional Mean Absolute Error}
\textbf{(Years)} \\
\midrule
\textbf{LifeSentence} & \textbf{61.6\% {[}61.4-61.8\%{]}} & \textbf{35.3\% {[}35.1-35.5\%{]}} & \textbf{1.15 {[}1.14-1.17{]}} \\
XGBoost & 58.2\% {[}58.0-58.5\%{]} & 15.0\% {[}14.8-15.1\%{]} & 1.78
{[}1.77-1.79{]} \\
LSTM & 56.4\% {[}56.1-56.6\%{]} & 11.6\% {[}11.5-11.8\%{]} & 2.30
{[}2.29-2.31{]} \\
GRU & 56.5\% {[}56.3-56.7\%{]} & 13.8\% {[}13.6-13.9\%{]} & 2.23
{[}2.21-2.24{]} \\
Logistic & 52.9\% {[}52.7-53.1\%{]} & 8.6\% {[}8.4-8.7\%{]} & 2.22
{[}2.21-2.23{]} \\
Transformer & 50.4\% {[}50.1-50.6\%{]} & 8.8\% {[}8.6-8.9\%{]} & 2.81
{[}2.80-2.83{]} \\
\bottomrule
\end{tabular}
\end{adjustbox}
\par}
\smallskip
{\small
\textbf{Table 1. Predictive performance on the next-event task.}
Comparison of LifeSentence against traditional (Logistic Regression,
XGBoost) and deep learning (LSTM, GRU, Transformer) baselines. Metrics
reported include Event Accuracy (correct prediction of event type),
Joint Accuracy (correct prediction of both event type and age), and
Conditional Mean Absolute Error (MAE, temporal error in years given a
correct event prediction). 95\% confidence intervals (calculated via
1,000 bootstrap samples) are provided in brackets.
\par}
\end{table}

Performance varies systematically across the life course (Supplementary
Fig. 1a--c). All models experience reduced accuracy during the 20--30
age range, consistent with characterizations of emerging adulthood as a
dense period of life events.\citeref{43--45} LifeSentence
maintains higher accuracy throughout this period, with joint accuracy
remaining above approximately 25\% even at peak volatility, whereas
baselines fall below 10\%. Performance is consistent across birth
cohorts from the 1930s through the 2000s (Supplementary Fig. 2d--f),
indicating that the model captures generalizable temporal dependencies
rather than cohort-specific artifacts. Analysis by event type reveals
that LifeSentence performs well on both socially normative structured
events (education completion) and more variable transitions (job
changes) (Supplementary Fig. 2a-c). Of particular note is mortality
prediction: LifeSentence predicts death as the next event with 29.8\%
accuracy, nearly threefold higher than the next-best baseline (XGBoost
at 10.3\%). When the model fails to predict death, it nearly always
predicts a medical diagnosis instead, which is contextually coherent
given that illness frequently precedes death in the observed data. This
error pattern is consistent with the model having learned the temporal
dependency structure between event types---specifically, the
illness--death sequence---rather than relying on marginal event
frequencies alone. Analysis of immediate next-event transition
probabilities confirms this: LifeSentence\textquotesingle s transition
probability error is threefold lower than all baselines (Supplementary
Fig. 3a--c). Baselines fail to learn, for example, that military service
start is almost always followed by service end. These one-step
transition results establish that the model has appropriately captured
shot-range event-transition dependencies.

\subsection*{Trajectory generation maintains structural accuracy over long horizons}

Next-event prediction assesses one step ahead local accuracy. To
determine whether LifeSentence maintains coherent narrative arcs over
longer horizons, we evaluated its capacity for complete trajectory
generation: predicting the full sequence of remaining life events from a
truncation point through the end of an individual\textquotesingle s
observed history. The combinatorial space of trajectories is
large.\citeref{47} With 20 event types across a typical 40-year
horizon, the space of possible trajectories is large; 96\% of
trajectories in our test set are unique by the eighth event
(Supplementary Fig. 5), meaning the model cannot rely on memorized
templates. On average, each test case requires predicting the next 23
years of an individual\textquotesingle s life.

Two illustrative cases are presented in Figure 2, one with a smaller
input history (truncated at age 18) and one with a more extensive input
history (truncated at age 29). For both, LifeSentence generates a
trajectory that closely mirrors the ground truth. Baseline models reveal
characteristic failures: the transformer degenerates into a dense
repetitive loop of events, while XGBoost recovers some of the correct
event types but at inappropriate time intervals.

\begin{figure}[tbp]
{\centering
\includegraphics[width=6.27014in,height=4.05139in]{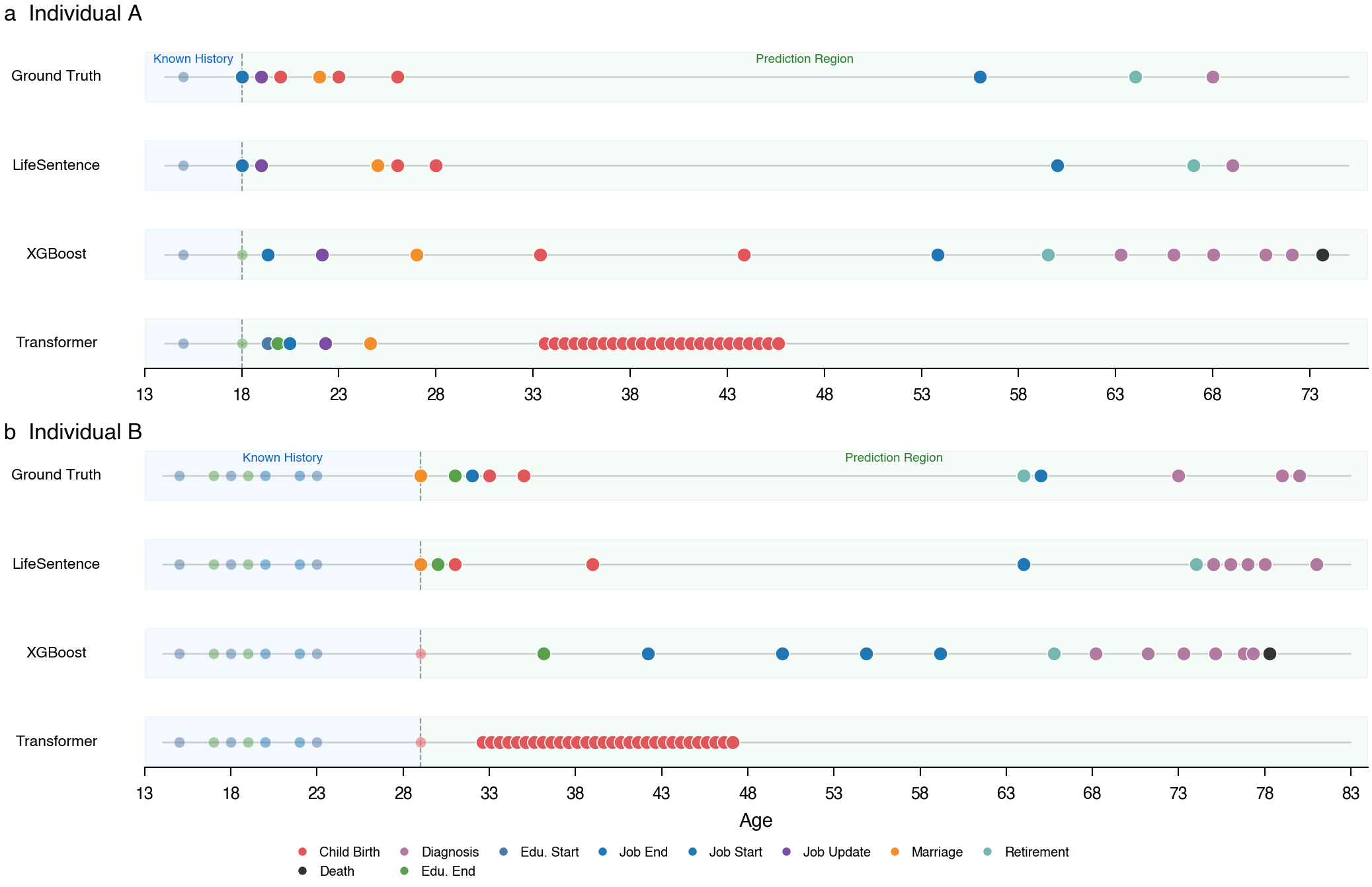}
\par}
\smallskip
{\small
\textbf{Figure 2}\,---\, \textbf{Trajectory generation across models
for two individuals}. Comparison of predicted life trajectories from
LifeSentence, XGBoost, and a from-scratch Transformer against ground
truth. Events prior to the truncation point (dashed line) constitute the
observed history (blue shading); events to the right are model
predictions (green shading) or ground-truth future events (top row of
each panel). (a) Individual A, truncated at age 18 with a prediction
horizon spanning approximately 55 years. LifeSentence produces a
trajectory that closely mirrors the ground truth in both event
composition and temporal spacing, capturing education completion,
marriage, childbirths, a long period of occupational stability,
retirement, and a late-life diagnosis. XGBoost recovers several correct
event types but distributes them unevenly. The from-scratch Transformer
exhibits mode collapse, degenerating into a dense repetitive loop of
events. (b) Individual B, truncated at age 29. LifeSentence again
recovers the major biographical arc with plausible temporal spacing. The
Transformer again collapses into repetitive early-life generation, while
XGBoost misses key events like marriage.
\par}
\end{figure}

Quantitative evaluation confirms this advantage across multiple
dimensions of trajectory accuracy (Supplementary Table 2). LifeSentence
achieves a Jaccard similarity\citeref{48} of 0.627 for
event-type overlap and a Levenshtein distance\citeref{49} of
8.10---meaning generated trajectories require roughly eight edits to
match reality, compared with 20 or more for deep learning baselines.
Full metric comparisons, including temporal displacement (Wasserstein
distance)\citeref{50} and event-frequency fidelity (Multiset
Jaccard)\citeref{51}, are reported in Supplementary Table 2.

LifeSentence also reproduces the non-uniform temporal distribution of
life events, accurately capturing elevated event density between ages 25
and 30 (Supplementary Fig. 6), while baseline models either do not
capture the appropriate peak or overestimate event frequency. Kernel
density estimates of age at childbirth, marriage, and death closely
track the empirical distributions, while baselines exhibit systematic
temporal misalignment (Fig. 3). Beyond aggregate distributional
fidelity, the model internalizes complex conditional probabilities:
female trajectories exhibit elevated probability of transitioning to
part-time employment following the birth of a child compared to men,
mirroring the ground-truth pattern\citeref{34,52} without
explicit supervision of this interaction (Supplementary Fig. 7). Direct
analysis of generative diversity confirms that LifeSentence produces a
substantially wider array of distinct biographical sequences and lower
rates of repetitive looping compared to baseline models (Supplementary
Fig. 8). Taken together, these results demonstrate that LifeSentence
generates diverse trajectories rather than collapsing to a modal
sequence.

\begin{figure}[tbp]
{\centering
\includegraphics[width=6.27014in,height=2.09028in]{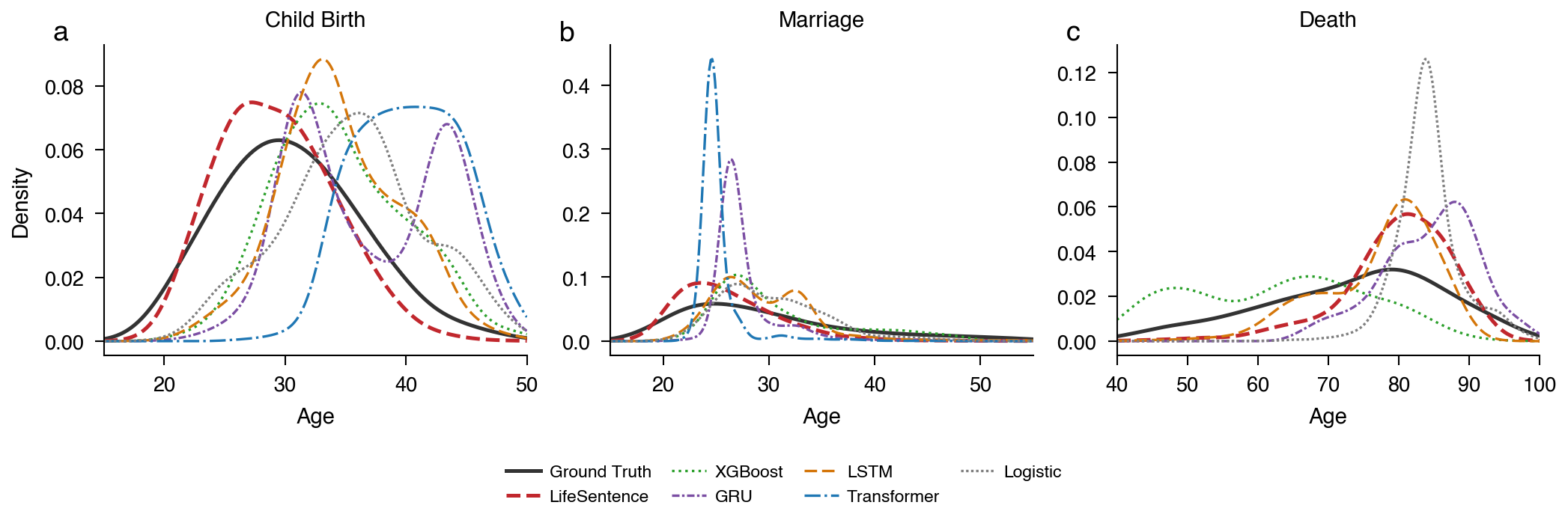}
\par}
\smallskip
{\small
\textbf{Figure 3\,---\, Macro-structural fidelity of generated life
events}. Kernel density estimation plots comparing the age-at-event
distributions for (a) Childbirth, (b) Marriage, and (c) Death across
Ground Truth, LifeSentence, and baseline models. LifeSentence closely
approximates the modal peaks and temporal width of the empirical
distributions, whereas other baselines exhibit temporal misalignment.
\par}
\end{figure}

Applied life-course research rarely calls for unconstrained event
generation. Rather, research questions are typically centered around
specific events (e.g., diagnosis)\citeref{53} or knowing what
happened between two observations years apart. These questions impose
temporal boundaries, type-specific targets, or structural endpoints that
require trajectories satisfying explicit constraints while maintaining
narrative coherence. Critically, constraints requiring semantic
comprehension ("predict until the next diagnosis," "connect this early
history to this late-life state") are difficult to express without a
natural-language interface.

We evaluated LifeSentence on six constrained generation tasks spanning
three constraint families: count and type-targeted generation (predict
the next three events, predict until a specific event, predict when a
specific event will happen), temporal-boundary generation (predict until
a specific year, predict what will happen in a certain year), and
endpoint-conditioned interpolation (generate trajectory between one
partial trajectory and a future state). Full metrics for all six tasks
are reported in Supplementary Tables 3a--b. On generation to a specified milestone task, LifeSentence satisfies
the termination constraint in 99.3\% of cases and achieves a Jaccard of
0.650, compared with the best baseline (LSTM) at 79.2\% constraint
satisfaction and 0.543 Jaccard (Supplementary Table 3b). Baselines
struggle because they lack any mechanism for conditioning generation on
a semantic target; they just generate until the target appears.
LifeSentence maintains a similar advantage on temporal-boundary
generation, achieving a Jaccard of 0.559 on the predict-up-to-year task
versus 0.441 for the best baseline.

The remaining three tasks require semantic comprehension of
natural-language instructions and are evaluable only for LifeSentence.
When asked to predict the timing of a specific future event type,
LifeSentence achieves a mean error of 4.23 years. When asked to generate
only events occurring during a specific future calendar year, it
achieves a Jaccard of 0.429. In the most structurally demanding task
(endpoint-conditioned interpolation), the model receives an
individual\textquotesingle s history up to age 25 and a single anchor
event around age 65, then generates the intervening four decades.
LifeSentence recovers this missing mid-life trajectory with a Jaccard of
0.587. An illustrative example (Supplementary Fig. 9) shows the model
receiving an individual\textquotesingle s history through age 25 and a
single endpoint (a diagnosis at age 66) and generating the intervening
four decades. The model can generate a mid-life sequence of career
entry, marriage, childbirths, and eventual retirement that is consistent
with both the early history and the late-life endpoint.

\subsection*{LifeSentence internalizes the temporal logic of the life course}

The preceding sections evaluated LifeSentence on forward-looking tasks:
predicting future events, generating trajectories, and satisfying
temporal or semantic constraints during generation. These results
establish that the model produces accurate forecasts, but they leave
open a distinct question: has the model acquired genuine structural
understanding of biographical logic, or is it exploiting surface-level
statistical regularities to generate plausible-looking
outputs.\citeref{54,55} To probe this, we designed tasks that
require no prediction of the future at all, only comprehension of the
past.

In the anomaly detection task, we insert a single foreign event randomly
drawn from the full event pool into an individual's trajectory, then ask
the model to identify it. LifeSentence identifies the anomaly with
93.9\% accuracy, demonstrating that it has learned not just what events
occur in human lives, but which events belong together within a single
life. Events embedded within tight institutional sequences like military
service and educational transitions are detected at near-perfect rates
because these events generate rigid contextual
expectations.\citeref{47} Relationship events such as divorce
and marriage prove hardest to detect, because their timing and
occurrence vary more substantially\citeref{56} across
individuals, so a foreign instance can plausibly pass as authentic
(Supplementary Fig. 4a). Detection accuracy also varies across the life
course (Supplementary Fig. 4b). The model performs best when anomalous
events are inserted before age 20 (97\%), and dips modestly during the
20--30 range (93\%).

The reordering task requires sorting a completely shuffled list of life
events into chronological order without access to any timestamps: the
model must rely entirely on its learned understanding of which events
logically precede others. Across 13,119 trajectories averaging 16.8
events in length, LifeSentence achieves a mean Kendall\textquotesingle s
Tau\citeref{57} of 91.2\%, meaning that the model places two
random events in the correct relative order over 90\% of the time. The
longest common subsequence\citeref{58} with the ground truth
spans 79.3\% of the trajectory on average, indicating that the order of
roughly four-fifths of the biography is recoverable without access to
any temporal information.

Together, these results indicate that LifeSentence has acquired an
internal representation of which life events belong together and in what
order they unfold. We next asked how LifeSentence responds when
biographical information is incomplete, and whether systematic removal
of events could reveal which ones carry the most structural weight. We
evaluated all models on next-event prediction under three masking
conditions of increasing severity: replacing individual events with
explicit mask tokens, silently deleting events without markers, and
redacting contiguous multi-year windows from the
history.\citeref{59--61} LifeSentence maintains the highest
accuracy across all three conditions (Supplementary Table 4).

Beyond aggregate robustness, these masking experiments reveal which
events carry the most structural weight. If the model has genuinely
learned which events belong together and in what order, then removing
structurally important events should degrade its predictions more than
removing peripheral ones.\citeref{62} We tested this by
measuring the change in next-event prediction performance by type and
timing of the masked event (Fig. 4).

\begin{figure}[tbp]
{\centering
\includegraphics[width=6.27014in,height=3.94028in]{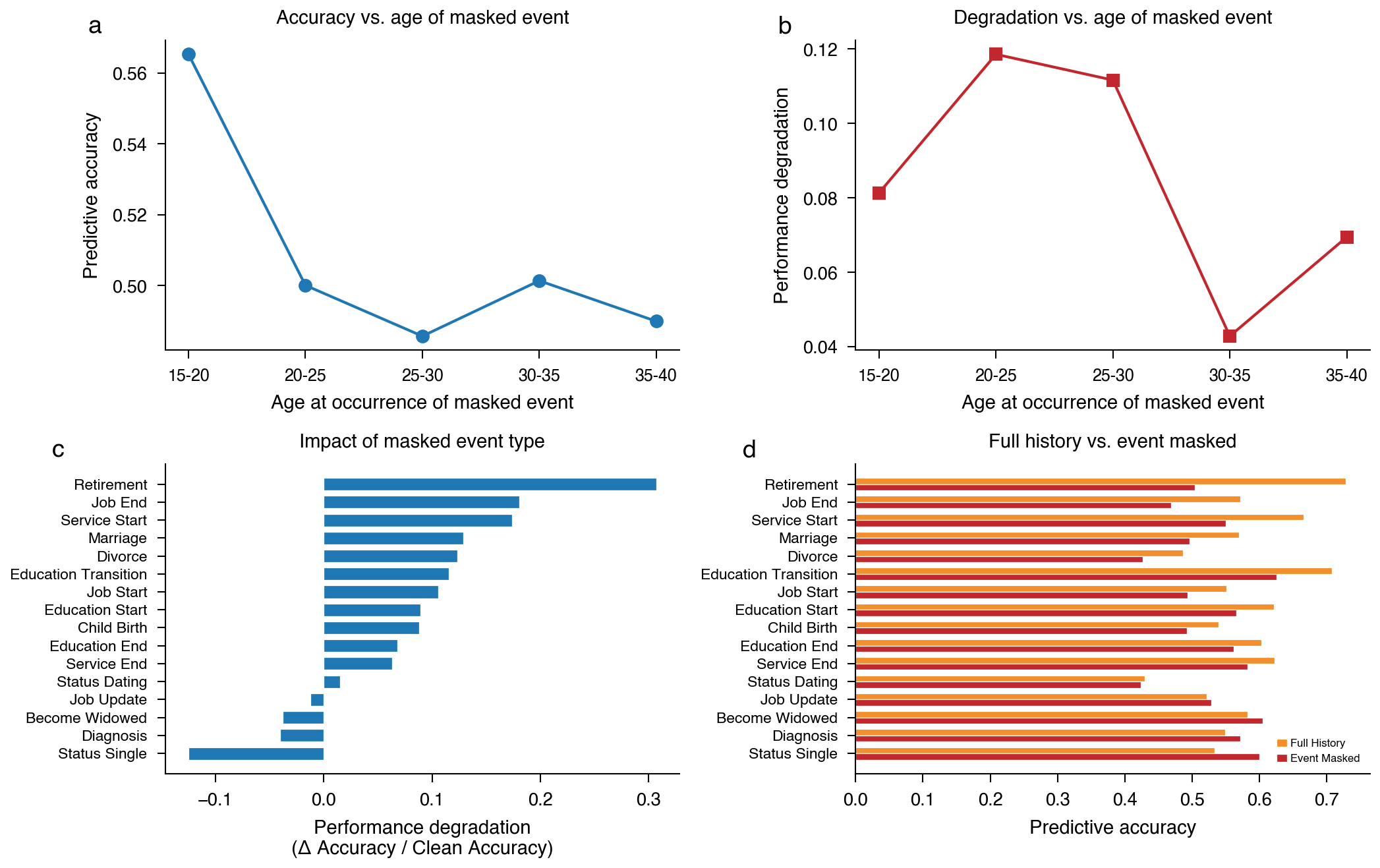}
\par}
\smallskip
{\small
\textbf{Figure 4\,---\, Sensitivity of predictive accuracy to
masking of historical events.} (a) Predictive accuracy on the next-event
task as a function of the age at which historical events are masked.
Accuracy is highest when events before age 20 are masked and drops to
its lowest at ages 25-30, indicating that events during this period
carry the most predictive information about an
individual\textquotesingle s future trajectory. (b) The same pattern
expressed as relative performance degradation (change in accuracy
normalised by clean-history accuracy). Degradation peaks at ages 20--25
(11.9\%). (c) Performance degradation by masked event type, showing the
normalised accuracy drop when each event type is removed from the
history. Masking retirement produces the largest degradation
(\textasciitilde0.30), followed by job end (\textasciitilde0.22),
service start (\textasciitilde0.20), and marriage (\textasciitilde0.18),
confirming that institutionally anchored transitions serve as
high-information structural pivots. (d) Absolute predictive accuracy
with full history (orange) versus with the event masked (red). The gap
between bars visualises the informational contribution of each event
type.
\par}
\end{figure}

Masking events in the 20--25 age range produces the largest accuracy
degradation, consistent with this period's role as a critical
developmental period\textsuperscript{43,44,45,{[}Roberts{]}}. The
specific type of missing information disproportionately affects
predictive capability: masking anchor events like military
service\citeref{63,64} start causes the highest degradation,
confirming that these institutionally structured transitions serve as
biographical anchors that constrain the space of possible futures.

We then tasked the model with imputing silently deleted events:
LifeSentence achieves 83.0\% accuracy for event-type identification, and
mean placement error of 3.5 years. Highly structured events (military
service completion, education completion) are recovered at rates above
0.95, while high-entropy transitions such as relationship status changes
prove difficult to impute (Supplementary Fig. 10). The pattern mirrors
the masking results: institutional events are both tightly constrained
by their surrounding context, making them recoverable when absent, and
serve as high-information structural pivots whose removal most disrupts
downstream prediction.

\subsection*{LifeSentence recovers known stratification patterns from event sequences}

Life events often co-occur with continuous outcomes central to social
science: income, occupational prestige, health, and life
satisfaction.\citeref{65--67} We trained LifeSentence on a
state imputation task in which the model receives an
individual\textquotesingle s discrete event trajectory, demographics
(age and gender), and must recover associated continuous measurements.
Crucially, the model receives no explicit supervision linking specific
covariates to specific outcomes: no encoded rule that men should earn
more, that college graduates should show steeper earnings growth, or
that childbirth should affect mothers\textquotesingle{} income
differently from fathers\textquotesingle.

Despite this, generated income trajectories recover known
sociodemographic stratification patterns. When stratified by education
(Fig. 5a), college-educated individuals show steeper income
growth\citeref{30,31} that accelerates through the 30s and 40s,
while those without college education plateau earlier and at lower
levels. The education premium in the model\textquotesingle s predictions
widens with age, consistent with the theory of cumulative advantage, in
which early educational investments compound into progressively larger
income differentials over the career. A similar pattern emerges when
stratifying by gender\citeref{32,33} (Supplementary Fig. 11).

\begin{figure}[tbp]
{\centering
\includegraphics[width=6.27014in,height=2.14375in]{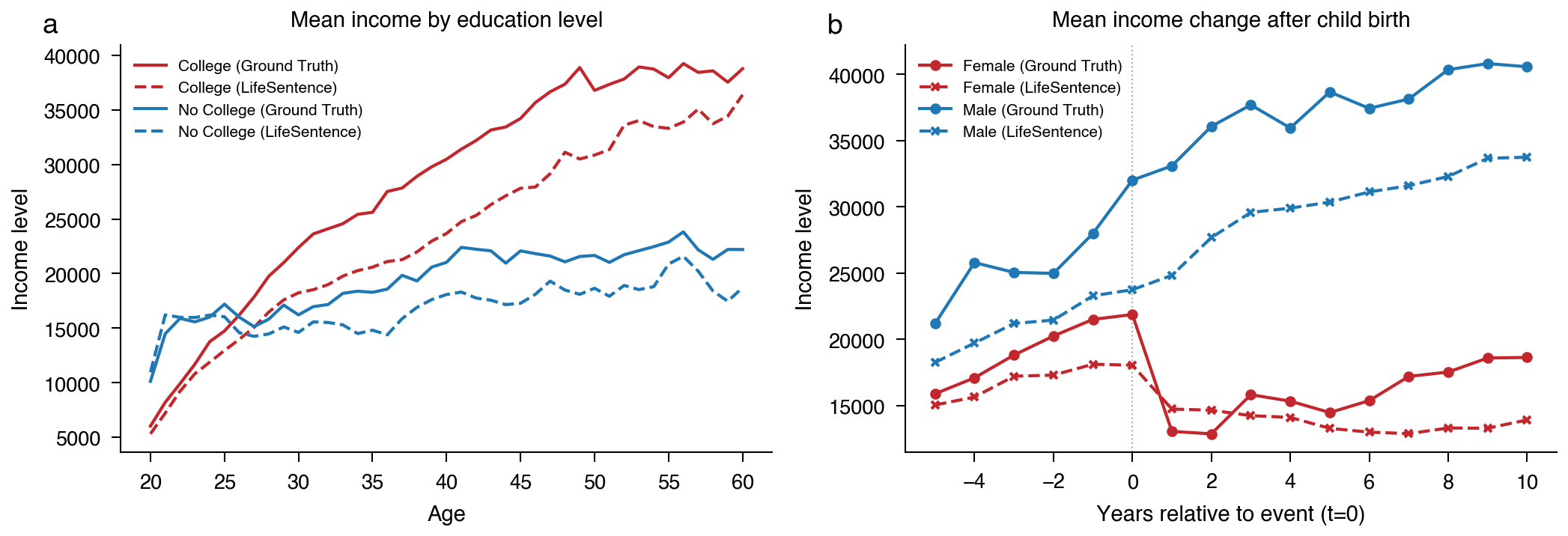}
\par}
\smallskip
{\small
\textbf{Figure 5\,---\, Sociodemographic stratification of income
trajectories.} (a) Mean predicted income level across the lifespan
stratified by education level, comparing college-educated (red) and
non-college-educated (blue) individuals. LifeSentence predictions
(dashed lines) recover the education premium without explicit
supervision: college-educated trajectories show steeper growth that
accelerates through the 30s and 40s, with the gap widening with age. (b)
Mean income level in the years surrounding the birth of a child (t = 0,
dotted vertical line) stratified by gender, comparing male (red) and
female (blue) individuals. The model recovers the "motherhood penalty":
male income continues to rise uninterrupted after childbirth, while
female income drops at birth and recovers only slowly over the
subsequent decade. Neither pattern was explicitly supervised (the model
was never provided rules linking education to earnings growth or
parenthood to gendered income divergence), suggesting that discrete
event sequences carry information about associated continuous outcomes.
\par}
\end{figure}

The model also recovers event-specific perturbations reflecting known
structural inequalities. Income trajectories anchored to childbirth
(Fig. 5b) reveal an asymmetry: male income continues rising
uninterrupted after childbirth, while female income drops sharply at the
moment of birth and recovers only slowly over the subsequent decade,
never fully closing the pre-birth gap. LifeSentence reproduces this
"motherhood penalty,"\citeref{34,35} capturing both the
immediate post-birth decline for women and the continued growth for men,
despite never being trained with any explicit representation of gendered
caregiving norms or labor market discrimination. For subjective
well-being, the model captures how diagnosis events produce immediate
declines in both life satisfaction and health
satisfaction,\citeref{68,69} tracking the slope and temporal
dynamics of these perturbations (Supplementary Fig. 11b).

Through exposure to the statistical co-occurrence of discrete event
sequences and continuous measurements across tens of thousands of
individuals, the model has learned that the biographical signatures
associated with gender, education, and parenthood systematically co-vary
with particular income and well-being trajectories. In effect, the
model's learned representations capture statistical associations between
event-sequence features and continuous outcomes that parallel documented
patterns of cumulative advantage, without these associations having been
specified as supervision targets.

\section*{Discussion}

LifeSentence demonstrates that pretrained language models can compensate
for smaller sample size in life-course prediction
tasks.\citeref{26,27,61,70,71} By encoding life events as
natural-language descriptions, we supplement a panel dataset with
distributional knowledge about education, employment, health, and
mortality acquired from trillions of tokens of pretraining
data.\citeref{72,73} The pretrained model already encodes
information about these concepts, while fine-tuning on rich biographical
panel data then teaches the model how these concepts interact across a
life: how an early educational decision shapes a career trajectory
decades later, or how a health event restructures the timing of
retirement. The divison of labor from pretraining and fine-tuning
reduces the sample size needed for competitive performance. A
from-scratch transformer trained on the same 65,000 individuals must
learn both concept semantics and biographical structure simultaneously.
The pretrained model's distributional representations provide a useful
initialization, and fine-tuning\citeref{74} need only adapt
these representations to life-course prediction and reasoning.

Beyond prediction, our results demonstrate that LifeSentence has
acquired structured representations of biographical logic. The anomaly
detection and reordering results demonstrate that LifeSentence has
internalized which events belong within a single biography and in what
order they unfold. The state imputation results go further, recovering
the education premium, gender income gap, and motherhood penalty without
explicit supervision. The masking experiments further reveal that this
information is not uniformly distributed: events during the 20--30 age
transition and institutionally anchored milestones carry more predictive
weight, a finding potentially relevant to longitudinal study design.

The natural-language interface extends the range of expressible tasks.
Prior architectures require task-specific output heads; LifeSentence
accepts free-text instructions, enabling constrained-generation queries
that would otherwise require bespoke model modifications.
LifeSentence\textquotesingle s capacity to interpret such instructions
extends the model from a prediction engine into an interactive research
tool.

Several limitations warrant consideration. First, LifeSentence was
trained exclusively on a single panel study. Whether learned
representations transfer to societies\citeref{75,76} with
fundamentally different institutional arrangements is an open question.
Second, the current 20-category event vocabulary compresses human
experience; major domains such as residential
mobility\citeref{77} and criminal justice\citeref{78}
involvement are absent. Third, the 24-billion-parameter model requires
4-bit quantization for feasible fine-tuning, representing a meaningful
infrastructure barrier. Future work should explore whether smaller
models\citeref{79} retain semantic transfer advantages. Fourth,
LifeSentence does not provide calibrated uncertainty estimates. The
model generates single trajectories rather than distributions over
possible futures, yet human lives are fundamentally stochastic.
Developing uncertainty quantification\citeref{80--82} is
critical before any individual-level application. Finally, the
pretrained language model\textquotesingle s semantic priors encode
societal biases from its training corpus. When LifeSentence recovers the
gender income gap, it is unclear to what extent these patterns reflect
empirical regularities in the data versus stereotypical associations
inherited from pretraining.\citeref{83--85}

These results indicate that instruction-tuning a pretrained language
model on structured biographical sequences can yield competitive
life-course prediction from panel-scale samples and point toward further
integration of distributional semantics with longitudinal social science
data.

\section*{Methods}

\subsection*{Ethics and Data Access}

The data analysis was conducted using the German Socio-Economic Panel
(SOEP),\citeref{36,37} a representative longitudinal household
survey administered by the German Institute for Economic Research (DIW
Berlin). Access is granted exclusively for scientific purposes and is
subject to strict confidentiality requirements that prohibit the
identification of individual respondents or the transmission of
microdata to unauthorized third parties. All model training and
inference were performed locally on university-managed infrastructure;
no individual-level data were transmitted to external APIs, cloud
services, or third-party model providers. The base language model
(Mistral Small 3.1) is an open-weight model whose parameters were
downloaded and fine-tuned entirely on-premises, ensuring that SOEP
microdata never left the secure research environment. Our model is a
research prototype intended for scientific investigation of life
trajectory predictability: before any deployment in applied contexts
such as policy planning or individual-level decision support, more
comprehensive auditing would be required to evaluate demographic
fairness across protected characteristics and to ensure appropriate
transparency regarding model limitations.

\subsection*{Dataset}

We constructed life-sequences from v37 of SOEP (1984-2020). The SOEP
records contains detailed information across multiple life domains for
each respondent, updated annually. We drew on five SOEP source files:
the biography module (biobirth, bioedu, biomarsy), the spell data module
(pbiospe), the individual questionnaire (pl, pkal), the individual-level
cross-national equivalent file (pequiv), and the person-level tracking
file (ppath, pgen), to assemble a comprehensive record of each
individual\textquotesingle s life trajectory.

We applied the following inclusion criteria: individuals must have valid
demographic attributes (sex coded as male or female, and a positive
birth year) and possess at least 10 discrete life events across the
observation period. The resulting filtered dataset contained 65,606
individuals. These individuals were randomly partitioned into training
(80\%, n = 52,485) and test (20\%, n = 13,121) sets at the individual
level, ensuring that no person appeared in both partitions and that all
tasks for a given individual were assigned entirely to one partition.

\subsubsection*{Employment data}

Employment records were extracted from the SOEP activity spell file
(pbiospe), which documents every discrete spell of biographical activity
for each respondent. Each spell is classified into one of ten
categories: school or college attendance, apprenticeship or vocational
training, military or community service, full-time employment, part-time
employment, unemployment, homemaker status, pensioner status, short-time
work, and observational gaps. For each spell, the dataset records the
start year and end year, yielding two events per spell (entry and exit).
Employment spells were further enriched with occupational information
from two additional sources. The biojob file provided occupational
classifications encoded via the International Standard Classification of
Occupations (ISCO-88), a hierarchical system that represents job types
with four-digit codes (for example, code 2111 references "Physicists and
Astronomers" and code 5141 references "Hairdressers"). The pgen file
provided the Standard International Occupational Prestige Scale (SIOPS),
a continuous index ranging from 6 to 78 that we discretized into five
ordinal categories: very low autonomy and unskilled work (6--32), low
autonomy and simple tasks (33--41), limited autonomy and intermediate
complexity (42--50), dependent or independent positions of moderate
prestige (51--63), and high-autonomy managerial positions (64+).
Positional seniority information from pgen further classified each
employment record by occupational rank (for example, unskilled worker,
skilled worker, civil servant, or self-employed). Retirement events were
identified from the calendar module (pkal) by flagging the first year in
which a respondent reported receiving a pension; individuals without an
early retirement record were assigned a statutory retirement event at
age 67.

\subsubsection*{Education data}

Education records were derived from the SOEP biography education module
(bioedu), which documents the full educational trajectory of each
respondent through nine milestone events: first and last attendance at
early childhood education and care, primary school enrollment,
transition to secondary school, exit from secondary school, entry into
vocational training, exit from vocational training, entry into tertiary
education, and exit from tertiary education. Each milestone records the
calendar year and, where applicable, the institution type and
certificate obtained. Vocational training types include apprenticeship,
specialized vocational school, health care school, specialized technical
school, and civil service training. Exit certificates include secondary
school degrees (secondary, intermediate, technical, upper secondary, and
dropout), vocational certificates (apprenticeship, vocational school,
technical school, civil service, and dropout), and tertiary degrees
(university of applied sciences, university, doctorate, and dropout).

\subsubsection*{Family data}

Family records were assembled from the biography marriage and
cohabitation spell file (biomarsy) and the biography birth file
(biobirth). The marriage spell file documents each period of a
respondent\textquotesingle s relationship status, with spell types
including: single, married, divorced, widowed, divorced or widowed
(unspecified), married but separated, and living in a registered
same-sex partnership. As with employment spells, each relationship spell
generates two events (start and end of the spell). The birth file
records the birth years of up to six children per respondent, yielding
child-birth events with birth-order information (first child, second
child, and so on).

\subsubsection*{Health data}

Health records were drawn from the individual-level cross-national
equivalent file (pequiv), which contains annual self-reported health
information. We extracted 19 binary health indicators spanning three
categories. Diagnostic conditions included stroke, hypertension,
diabetes, cancer, psychiatric problems, arthritis, angina or heart
conditions, and asthma or breathing difficulties (8 indicators).
Functional limitations included difficulty climbing stairs, bathing,
dressing, getting out of bed, shopping, walking, doing housework, and
bending or lifting, as well as a general indicator that health limits
vigorous physical activities (9 indicators). A hospitalization indicator
recorded whether an accident required hospitalization, and a disability
indicator recorded whether the respondent had become disabled. Mortality
information was extracted from the person-level tracking file (ppath),
which records the year of death for deceased respondents.

\subsubsection*{State variables}

In addition to discrete life events, we recorded four continuous state
variables measured annually: individual gross income (from pequiv),
overall life satisfaction (from pequiv, on a 0--10 scale), health
satisfaction (from pequiv, on a 0--10 scale), and occupational prestige
(SIOPS score from pgen, discretized into five ordinal categories as
described above). Unlike discrete events, these variables represent
periodic measurements that characterize an individual\textquotesingle s
circumstances at particular time points. Negative or missing values were
set to missing prior to analysis. Income values were retained as
continuous measurements.

\subsection*{Sequence Construction and Preprocessing}

From each appropriate SOEP source file, events were assigned one of
the following 20 discrete event types:

{\small\renewcommand{\arraystretch}{1.05}%
\begin{center}
\begin{tabular}{@{}llll@{}}
\texttt{JOB\_START} & \texttt{EDUCATION\_START} & \texttt{CHILD\_BIRTH} & \texttt{BECOME\_WIDOWED} \\
\texttt{JOB\_END} & \texttt{EDUCATION\_END} & \texttt{MARRIAGE} & \texttt{STATUS\_SINGLE} \\
\texttt{JOB\_UPDATE} & \texttt{EDUCATION\_TRANSITION} & \texttt{DIVORCE} & \texttt{STATUS\_DATING} \\
\texttt{RETIREMENT} & \texttt{SERVICE\_START} & \texttt{SEPARATION} & \texttt{PARTNER\_ABROAD} \\
\texttt{DIAGNOSIS} & \texttt{SERVICE\_END} & \texttt{REMARRIAGE} & \texttt{DEATH} \\
\end{tabular}
\end{center}} Each event carries associated detail fields
specific to its domain (for example, occupation name, education level,
diagnosis, or child birth order), the individual\textquotesingle s age
at the time of the event, and the calendar year.

We applied several refinement procedures to the raw event sequences.
First, sequential medical diagnoses occurring within the same calendar
year were aggregated into a single DIAGNOSIS event containing a
deduplicated list of conditions, reducing redundancy while preserving
diagnostic information. Second, employment events were contextualized
based on persistent employment status: a JOB\_START event occurring
while an individual was already in continuous employment (full-time or
part-time) was reclassified as JOB\_UPDATE to distinguish between
initial labor market entry and job changes within an unbroken employment
spell. Third, within each calendar year, events were sorted so that
DEATH always appeared last, ensuring that the temporal ordering of
within-year events respected logical constraints. Finally, consecutive
duplicate entries were suppressed: for example, if a
respondent\textquotesingle s relationship status in year t was identical
to the status already recorded in the most recent prior year, the
redundant entry was removed.

\subsubsection*{Event representation}

Each life event was represented not as a categorical index but as a
structured natural-language record. Employment events included
occupation names drawn from ISCO-88 descriptors (for example,
"Mechanical Engineers," "Secretaries"), positional seniority labels, and
employment type. Education events included institution names (for
example, "Upper secondary school"), certificate types, and educational
levels drawn from the German system\textquotesingle s standard
terminology. Health events included diagnostic condition names. Family
events included relationship status labels and child birth order.

\subsubsection*{Prompt template design}

Each JSONL training example was converted into a structured
natural-language prompt for supervised fine-tuning of the language
model. Specific details of prompt formatting and layout are detailed in
the Supplementary Information.

\subsection*{Task Formulation}

A single prediction task cannot determine whether a model has genuinely
learned the structure of human life trajectories or is merely exploiting
surface-level correlations. To enable comprehensive evaluation, we
formulated an 18-task taxonomy organized into three families of
increasing inferential demand, each targeting a different dimension of
model capability. Forecasting tasks assess whether the model can
accurately predict future events at varying horizons, from single
next-event prediction to complete remaining-life trajectory generation.
Robustness tasks test whether predictions remain reliable when
biographical information is incomplete, simulating the missing data
conditions common in longitudinal panel studies. Reasoning tasks probe
whether the model has acquired genuine structural understanding of
biographical logic (such as the ability to detect anomalous events,
recover chronological order without timestamps, and translate discrete
event sequences into the continuous socioeconomic trajectories they
produce). Together, these three families evaluate not only predictive
accuracy but also the model's capacity to internalize meaningful
biographical structure. The 18-task taxonomy comprises the following
categories:

\paragraph{Forecasting tasks.}

\begin{itemize}
\item
  NEXT\_EVENT: given a prefix of the event history and the corresponding
  state history, predict the immediate next discrete life event; this
  task was generated exhaustively using a sliding window over the entire
  trajectory, producing one example for every valid split point.
\item
  NEXT\_3\_EVENTS: predict the next three events in chronological order.
\item
  PREDICT\_UNTIL\_EVENT: predict all future events up to and including
  the next occurrence of a specified event.
\item
  ENTIRE\_TRAJECTORY: predict the complete remaining life trajectory
  from the current point.
\item
  PREDICT\_UP\_TO\_YEAR: predict all events occurring up to a specified
  future year.
\item
  SNAPSHOT\_YEAR: predict the specific events occurring in a randomly
  selected future year.
\item
  NEXT\_SPECIFIC\_TYPE: predict the details of the next occurrence of a
  target event type.
\end{itemize}

\paragraph{Robustness tasks.} These tasks tested model resilience to
incomplete information.

\begin{itemize}
\item
  MASKED\_HISTORY\_PREDICT\_FUTURE and
  MASKED\_HISTORY\_PREDICT\_NEXT\_EVENT: up to two events in the history
  were replaced with explicit {[}MASKED EVENT{]} tokens, and the
  corresponding state history entries for those years were removed.
\item
  IMPLICIT\_GAP\_PREDICT\_FUTURE and
  IMPLICIT\_GAP\_PREDICT\_NEXT\_EVENT: up to two events were silently
  removed from the history without any mask token, and state entries for
  the affected years were also removed.
\item
  MASKED\_PERIOD\_PREDICT\_FUTURE and
  MASKED\_PERIOD\_PREDICT\_NEXT\_EVENT: a contiguous block of 5--10
  years was redacted from the history, with all events in that range
  replaced by a descriptive mask token indicating the redacted period,
  and state entries within that range removed.
\end{itemize}

\paragraph{Reasoning tasks.}

\begin{itemize}
\item
  CONNECT\_POINTS: given the history up to approximately age 25 and a
  single target event at approximately age 65 (along with state history
  up to the start point), generate the plausible sequence of events
  connecting these two points.
\item
  IMPLICIT\_REMOVAL: given a timeline with one silently removed event,
  identify the missing event and its correct position.
\item
  ANOMALY\_DETECTION: given a timeline with one injected distractor
  event (sampled from the training event pool with temporally consistent
  age and year interpolation), identify the anomalous event.
\item
  STATE\_IMPUTATION: given the discrete event history alone (no state
  variables), estimate the full continuous state history.
\item
  REORDERING: given a shuffled set of events with age and year fields
  removed, reconstruct the correct chronological order.
\end{itemize}

All tasks except NEXT\_EVENT were generated at up to three randomly
sampled split points per individual to control dataset size, while
NEXT\_EVENT was generated exhaustively at every valid split point. The
CONNECT\_POINTS, IMPLICIT\_REMOVAL, ANOMALY\_DETECTION,
STATE\_IMPUTATION and REORDERING tasks were each generated once per
individual from the full trajectory.

\subsection*{Model Architecture}

\subsubsection*{LifeSentence: Large Language Model Approach}

Our primary model, LifeSentence, adapts a pretrained large language
model to life trajectory prediction through supervised fine-tuning. We
employ Low-Rank Adaptation (LoRA)\citeref{40} to enable
efficient parameter-efficient training while preserving the pretrained
model\textquotesingle s capabilities. Each training example was
converted into the natural-language prompt format described (see
Supplementary Information \emph{Sequence Construction and
Preprocessing}), and the fine-tuned model was trained to produce the
corresponding JSON target via supervised next-token prediction.

\paragraph{Model architecture and training.} We fine-tuned Mistral Small
3.1 24B Base (unsloth/Mistral-Small-3.1-24B-Base-2503), a
24-billion-parameter decoder-only language model, using the Unsloth
library for efficient training and inference. Full finetuning of a
24B-parameter model exceeds available GPU memory; we therefore adopt the
QLoRA\citeref{86} framework, where the frozen base model weights
were stored in 4-bit NormalFloat quantization, while the trainable
adapter parameters and all gradient computations used bfloat16 precision
throughout. Low-Rank Adaptation (LoRA) was applied with rank r\,=\,16 and
$\alpha$\,=\,16 (effective scaling factor of 1.0), targeting all attention and
feed-forward projection matrices. No dropout was applied to the LoRA
layers, and no bias terms were trained.

Training was conducted for 1 epoch on a single NVIDIA A100 with the
following hyperparameters: learning rate of 2\,$\times$\,10\textsuperscript{-5};
paged AdamW optimizer (32-bit);\citeref{87} linear learning-rate
scheduler with a 3\% warmup ratio;\citeref{88} per-device batch
size of 8; gradient accumulation over 4 steps (effective batch size of
32);\citeref{89} maximum gradient norm of 0.3.

\paragraph{Inference and evaluation.} At inference time, the fine-tuned
model was loaded from the saved checkpoint in 4-bit quantization using
Unsloth's FastLanguageModel and placed in inference mode. Prompts were
constructed identically to training using the same formatting functions.
All predictions were generated using greedy decoding (temperature = 0).
Model outputs were parsed by locating the outermost pair of curly braces
in the generated text and attempting JSON deserialization of the
enclosed substring. Outputs that failed to parse were retained in the
evaluation dataset and scored as incorrect predictions across all
metrics; no outputs were excluded from reported results.

\subsection*{Benchmark Models}

We benchmarked LifeSentence against two classes of baseline models that
represent the dominant approaches in the life-course prediction
literature. Classical machine learning models (logistic regression and
XGBoost) operate on hand-engineered fixed-length feature vectors
extracted from the event history, discarding sequential ordering but
providing strong baselines on tabular prediction tasks. Deep learning
models (Transformer encoder, LSTM, and GRU) receive the raw event
sequence as input and can in principle learn temporal dependencies
directly from data, but must learn both event semantics and sequential
structure from scratch using only the panel data. This two-class design
allows us to disentangle the contributions of sequential modeling (deep
learning vs. classical) from the contribution of pretrained semantic
knowledge (LifeSentence vs. all baselines). All baselines were trained
and evaluated on the same data partitions and task definitions as
LifeSentence.

\subsubsection*{Classical machine learning baselines}

\paragraph{Feature extraction.} All classical baselines shared a common
feature extraction pipeline. For each example, a fixed-length feature
vector was computed from the event history, state history and
demographic information. Demographic features comprised binary
indicators for gender, normalized birth year, and normalized birth
decade. Event count features included, for each of the 20 event types,
the total count and a binary has-occurred indicator, and for each of the
5 event domains (family, education, employment, health, service), the
total domain count. Temporal features comprised normalized current age,
minimum observed age, age span, total number of events (raw and
log-transformed), year span, latest observed year and one-hot encoded
age bins (0--18, 18--25, 25--35, 35--45, 45--55, 55--65, 65--100).
Last-event features included one-hot encoding of the most recent event's
type and domain, normalized age of the last event and time elapsed since
the last event. Bigram features captured transition patterns via one-hot
encoding of the second-to-last event type. State features comprised the
most recent income (normalized by dividing by 100,000, plus
log-transformed), life satisfaction score and health satisfaction score
(each normalized to 0--1 by dividing by 10). Recent context features
captured the proportional frequency of each event type within the last 5
events. All features were standardized using a StandardScaler fitted on
the training set; NaN and infinite values were replaced with zero prior
to scaling.

\paragraph{Task categorization.} The multi-task evaluation set was
partitioned into three categories with respect to classical model
capabilities. Single-event prediction tasks (NEXT\_EVENT,
MASKED\_HISTORY\_PREDICT\_NEXT\_EVENT,
IMPLICIT\_GAP\_PREDICT\_NEXT\_EVENT,
MASKED\_PERIOD\_PREDICT\_NEXT\_EVENT) were evaluated directly. Sequence
generation tasks (NEXT\_3\_EVENTS, PREDICT\_UNTIL\_EVENT,
ENTIRE\_TRAJECTORY, PREDICT\_UP\_TO\_YEAR, and their masked/corrupted
variants) were handled through autoregressive rollout. Seven task types
(NEXT\_SPECIFIC\_TYPE, SNAPSHOT\_YEAR, CONNECT\_POINTS,
IMPLICIT\_REMOVAL, ANOMALY\_DETECTION, REORDERING, STATE\_IMPUTATION)
were deemed unsupported because they required capabilities beyond point
prediction and were excluded from classical model evaluation.

\paragraph{XGBoost.} The XGBoost\citeref{90} baseline employed two
models: an XGBClassifier for event type prediction and an XGBRegressor
for event age prediction. The classifier was configured with the
multi:softmax objective, 100 estimators, maximum depth 6, learning rate
0.1, and histogram-based tree construction. The regressor used 100
estimators, maximum depth 5 and learning rate 0.1, and was trained only
on examples with valid target ages. We trained on all single-event
prediction tasks including masked and corrupted variants, and ``all''
(every task yielding a valid target event type). For sequence generation
tasks, the model was applied autoregressively: after predicting the next
event type and age, a synthetic event dictionary was appended to the
running history, features were re-extracted from the updated history,
and the next prediction was generated. Stopping criteria included death
prediction, reaching a specified target year or reaching a task-specific
maximum sequence length (3 for NEXT\_3\_EVENTS, 30 for bounded tasks, 50
for full trajectory prediction).

\paragraph{Logistic regression.} The logistic regression baseline followed
an identical pipeline to XGBoost, substituting the tree-based models
with linear models. A multinomial logistic regression classifier (L-BFGS
solver, maximum 1,000 iterations) was used for event type prediction,
and a standard linear regression model was used for age prediction. Both
models used the same StandardScaler and feature extraction pipeline. The
same training mode, autoregressive sequence generation procedure and
stopping criteria as XGBoost were employed.

\subsubsection*{Deep learning models}

Three deep learning models---a Transformer encoder,\citeref{17}
an LSTM\citeref{91} and a GRU\citeref{92}---shared a
common base architecture. The input representation consisted of four
fused embedding components, summed element-wise: (1) a learned
event-type embedding of dimension d\,=\,256 for each token in the
vocabulary (20 event types plus 4 special tokens:
\textless PAD\textgreater, \textless SOS\textgreater,
\textless EOS\textgreater, \textless UNK\textgreater); (2) a linear
projection of the normalized age (age/100) to d dimensions; (3) a
learned gender embedding broadcast across the sequence length; and (4) a
linear projection of the normalized birth year
((birth\_year\,-\,1900)/100) broadcast across the sequence. These four
components were passed through a dropout layer (rate 0.1) to produce the
fused input representation. Each model produced two outputs from the
final hidden state: event type logits via a linear classification head
mapping d to the vocabulary size, and a normalized predicted age via a
single-output linear regression head. The training objective combined
cross-entropy loss for event type classification with mean squared error
loss for age prediction.

\paragraph{Transformer.} The Transformer model used three
TransformerEncoderLayers with d\,=\,256, 4 attention heads and dropout
rate 0.1. The prediction was derived from the hidden state of the last
valid (non-padding) token in each sequence. Positional encoding as well
as a causal mask are applied. Training used the Adam optimizer with a
learning rate of 1\,$\times$\,10\textsuperscript{-3}, batch size 128 and up to 30
training epochs, with early stopping after 15
epochs.\citeref{93}

\paragraph{LSTM and GRU.} The LSTM and GRU models used the same input
fusion and output heads. The LSTM employed a three-layer LSTM with
hidden dimension 256. The GRU used an identically configured three-layer
GRU. Both models leveraged PyTorch's efficient handling of
variable-length inputs, using the padding mask to compute true sequence
lengths. The final hidden state (h$_n$) was used as the summary vector for
prediction. Training used the Adam optimizer with learning rate
1\,$\times$\,10\textsuperscript{-3}, batch size 128 and up to 30 training epochs,
with early stopping after 15 epochs.

\paragraph{Inference.} At test time, all three neural models used greedy
decoding: the event type with the highest logit was selected, and the
predicted normalized age was rescaled (multiplied by 100) and clamped to
be at least 0.5 greater than the last observed age to enforce
monotonicity. For sequence generation tasks, predictions were generated
autoregressively by appending each predicted event to the running
history and re-encoding the updated sequence, up to task-specific
maximum step counts (1 for single-event tasks, 3 for NEXT\_3\_EVENTS, 30
for other sequence tasks). Generation terminated upon predicting DEATH,
\textless EOS\textgreater, \textless PAD\textgreater{} or exceeding age
100.

\subsection*{Evaluation Metrics}

We employed a list of metrics to assess model performance across the
18-task taxonomy, as no single metric captures all dimensions of
life-course prediction. Each metric is designed to evaluate a specific
aspect of model capability. Event-level classification metrics (Event
Type Accuracy, Joint Accuracy, Conditional MAE) assess single-event
forecasting performance and are the primary measures for prediction
tasks. Sequence-level trajectory metrics (Jaccard Similarity, Multiset
Jaccard, Levenshtein Distance, Wasserstein Distance) evaluate the
structural and temporal fidelity of generated multi-event trajectories,
serving as the primary measures for trajectory generation and
constrained generation tasks. Specialised classification metrics
(Anomaly Detection Accuracy, Imputation Accuracy) and ordering metrics
(Kendall's $\tau$, Longest Common Subsequence) evaluate the reasoning tasks
that probe whether the model has acquired genuine structural
understanding of biographical logic. Finally, constraint satisfaction
measures assess whether generated trajectories comply with task-specific
requirements, providing a complementary evaluation of the model's
ability to follow natural-language instructions. Metrics are organised
below by these categories; full results are reported in the
corresponding tables and figures.

\paragraph{Event-level classification metrics.} Event Type Accuracy
measures the proportion of predictions where the predicted event type
matches the ground truth. Joint Accuracy requires both a correct event
type and a predicted age to be correct (after rounding). Conditional
Mean Absolute Error (Conditional MAE) reports the mean absolute age
error in years, computed only over predictions where the event type was
correctly identified. These metrics are reported for single-event tasks
(Table 1, Supplementary Fig. 1, Supplementary Fig. 2) and for the
next-three-events task (Supplementary Table 3a), where they are computed
independently per slot and averaged across positions.

\paragraph{Specialised classification metrics.} Anomaly Detection Accuracy
measures the proportion of trajectories in which the model correctly
identifies an injected distractor event (Supplementary Fig. 4).
Imputation Accuracy measures the proportion of silently deleted events
whose type is correctly recovered from surrounding context; Placement
Accuracy further requires that the recovered event be assigned to the
correct temporal position (Supplementary Fig. 10).

\paragraph{Sequence-level trajectory metrics.} For trajectory generation
tasks, we report four complementary metrics. Jaccard
Similarity\citeref{48} measures the set overlap between
predicted and ground-truth unique event types. Multiset
Jaccard\citeref{51} extends this to account for event-type
frequency (e.g., penalising a trajectory that predicts two childbirths
when three occurred). Levenshtein Distance\citeref{49} measures
the minimum number of insertions, deletions, and substitutions required
to transform the predicted event-type sequence into the ground truth.
Wasserstein Distance\citeref{50} measures the optimal transport
cost of moving predicted event ages to their ground-truth temporal
positions. These are reported in Supplementary Table 2 (unconstrained
generation) and Supplementary Table 3b (constrained generation).

\paragraph{Constraint satisfaction.} For the PREDICT\_UNTIL\_EVENT task,
Constraint Satisfaction Rate measures the proportion of generated
trajectories that actually terminate at the requested milestone event
(Supplementary Table 3b).

\textbf{Ordering and structural metrics.} For the reordering task, we
report Kendall\textquotesingle s $\tau$\citeref{57}, which measures
the proportion of concordant event pairs between predicted and true
orderings, and Longest Common Subsequence\citeref{58}
proportion, which captures the fraction of the trajectory recovered as
an unbroken in-order chain. Transition probability error quantifies the
absolute difference between the model\textquotesingle s predicted
pairwise event-transition probabilities and ground-truth transition
matrices (Supplementary Fig. 3).

\textbf{Data availability}

The data used in this study are not publicly available due to German
Data Protection laws. Access to the data can be obtained via the German
Institute for Economic Research (DIW).

\textbf{Code availability}

The source code for the data pre-processing, model training, analysis
and visualization is available on GitHub at
https://github.com/samuelliu2019/life-sentence.

\section*{References}
\vspace{-0.4ex}

{\footnotesize
\setlength{\parskip}{1pt plus 1pt minus 0pt}%
\interlinepenalty=0%
\begin{enumerate}[itemsep=1.5pt,parsep=0pt,topsep=1pt,leftmargin=*,labelsep=0.5em]
\def\labelenumi{\arabic{enumi}.}
\item
  Vaillant, G. E. \& Mukamal, K. Successful aging. Am. J. Psychiatry
  158, 839--847 (2001).
\item
  Diener, E. \& Seligman, M. E. P. Very happy people. Psychol. Sci. 13,
  81--84 (2002).
\item
  Moffitt, T. E. et al. A gradient of childhood self-control predicts
  health, wealth, and public safety. Proc. Natl Acad. Sci. USA 108,
  2693--2698 (2011).
\item
  Salganik, M. J. et al. Measuring the predictability of life outcomes
  with a scientific mass collaboration. Proc. Natl Acad. Sci. USA 117,
  8398--8403 (2020).
\item
  Grossmann, I. et al. Insights into the accuracy of social
  scientists\textquotesingle{} forecasts of societal change. Nat. Hum.
  Behav. 7, 484--501 (2023).
\item
  Oparina, E. et al. Machine learning in the prediction of human
  wellbeing. Sci. Rep. 15, 1632 (2025).
\item
  Halpern-Manners, A., Warren, J. R., Raymo, J. M. \& Nicholson, D. A.
  The impact of work and family life histories on economic well-being at
  older ages. Social Forces 93, 1369--1396 (2015).
\item
  Wills, A. K. et al. Life course trajectories of systolic blood
  pressure using longitudinal data from eight UK cohorts. PLoS Medicine
  8, e1000440 (2011).
\item
  Hayward, M. D. \& Gorman, B. K. The long arm of childhood: the
  influence of early-life social conditions on men\textquotesingle s
  mortality. Demography 41, 87--107 (2004).
\item
  Ferraro, K. F., Schafer, M. H. \& Wilkinson, L. R. Childhood
  disadvantage and health problems in middle and later life: early
  imprints on physical health? American Sociological Review 81, 107--133
  (2016).
\item
  Willson, A. E., Shuey, K. M. \& Elder, G. H. Jr. Cumulative advantage
  processes as mechanisms of inequality in life course health. American
  Journal of Sociology 112, 1886--1924 (2007).
\item
  Rook, K. S., Catalano, R. \& Dooley, D. The timing of major life
  events: effects of departing from the social clock. American Journal
  of Community Psychology 17, 233--258 (1989).
\item
  Elder, G. H., Jr. The life course as developmental theory. Child Dev.
  69, 1--12 (1998).
\item
  Elder, G. H., Jr. \& Shanahan, M. J. The life course and human
  development. In Handbook of Child Psychology Vol. 1 (ed. Lerner, R.
  M.) 665--715 (Wiley, 2006).
\item
  Topel, R. H. \& Ward, M. P. Job mobility and the careers of young men.
  Q. J. Econ. 107, 439--479 (1992).
\item
  Lynch, J. W., Kaplan, G. A. \& Shema, S. J. Cumulative impact of
  sustained economic hardship on physical, cognitive, psychological, and
  social functioning. N. Engl. J. Med. 337, 1889--1895 (1997).
\item
  Vaswani, A. et al. Attention is all you need. In Advances in Neural
  Information Processing Systems 30 (NeurIPS) (2017).
\item
  Devlin, J., Chang, M.-W., Lee, K. \& Toutanova, K. BERT: Pre-training
  of deep bidirectional transformers for language understanding. In
  Proc. NAACL-HLT 2019, 4171--4186 (2019).
\item
  Shmatko, A. et al. Learning the natural history of human disease with
  generative transformers. Nature 647, 248--256 (2025).
\item
  Renc, P. et al. Zero shot health trajectory prediction using
  transformer. npj Digit. Med. 7, 256 (2024).
\item
  Vafa, K., Palikot, E., Du, T., Kanodia, A., Athey, S. \& Blei, D. M.
  CAREER: a foundation model for labor sequence data. Preprint at
  https://arxiv.org/abs/2202.08370 (2024).
\item
  Savcisens, G. et al. Using sequences of life-events to predict human
  lives. Nat. Comput. Sci. 4, 43--56 (2024).
\item
  Yang, E., Li, M. D., Raghavan, S., Deng, F., Lang, M., Succi, M. D.,
  Huang, A. J. \& Kalpathy-Cramer, J. Transformer versus traditional
  natural language processing: how much data is enough for automated
  radiology report classification? Br. J. Radiol. 96, 20220769 (2023).
\item
  Grinsztajn, L., Oyallon, E. \& Varoquaux, G. Why do tree-based models
  still outperform deep learning on typical tabular data? in Advances in
  Neural Information Processing Systems 35 (NeurIPS, 2022).
\item
  Raffel, C. et al. Exploring the limits of transfer learning with a
  unified text-to-text transfer transformer. J. Mach. Learn. Res. 21,
  1--67 (2020).
\item
  Wei, J. et al. Finetuned language models are zero-shot learners. In
  Proc. International Conference on Learning Representations (2022).
\item
  Du, T., Kanodia, A., Brunborg, H., Vafa, K. \& Athey, S. LABOR-LLM:
  language-based occupational representations with large language
  models. Preprint at https://arxiv.org/abs/2406.17972 (2024).
\item
  Aribandi, V. et al. ExT5: towards extreme multi-task scaling for
  transfer learning. In Proc. International Conference on Learning
  Representations (2022).
\item
  Lundberg, I., Brown-Weinstock, R., Clampet-Lundquist, S., Pachman, S.,
  Nelson, T. J., Yang, V., Edin, K. \& Salganik, M. J. The origins of
  unpredictability in life outcome prediction tasks. Proc. Natl Acad.
  Sci. USA 121, e2322973121 (2024).
\item
  Tamborini, C. R., Kim, C. \& Sakamoto, A. Education and lifetime
  earnings in the United States. Demography 52, 1383--1407 (2015).
\item
  Bhuller, M., Mogstad, M. \& Salvanes, K. G. Life-cycle earnings,
  education premiums, and internal rates of return. J. Labor Econ. 35,
  993--1030 (2017).
\item
  Bertrand, M., Goldin, C. \& Katz, L. F. Dynamics of the gender gap for
  young professionals in the financial and corporate sectors. Am. Econ.
  J. Appl. Econ. 2, 228--255 (2010).
\item
  Barth, E., Kerr, S. P. \& Olivetti, C. The dynamics of gender earnings
  differentials: evidence from establishment data. Eur. Econ. Rev. 134,
  103713 (2021).
\item
  Kleven, H., Landais, C. \& Søgaard, J. E. Children and gender
  inequality: evidence from Denmark. Am. Econ. J. Appl. Econ. 11,
  181--209 (2019).
\item
  Almond, D., Cheng, Y. \& Machado, C. Large motherhood penalties in US
  administrative microdata. Proc. Natl Acad. Sci. USA 120, e2209740120
  (2023).
\item
  Goebel, J. et al. The German Socio-Economic Panel (SOEP). Jahrb.
  Natl.ökon. Stat. 239, 345--360 (2019).
\item
  Wagner, G. G., Frick, J. R. \& Schupp, J. The German Socio-Economic
  Panel Study (SOEP) -- scope, evolution and enhancements. Schmollers
  Jahrb. 127, 139--169 (2007).
\item
  Gurnee, W. \& Tegmark, M. Language models represent space and time. In
  Proc. International Conference on Learning Representations (2024).
\item
  Aghajanyan, A., Gupta, S. \& Zettlemoyer, L. Intrinsic dimensionality
  explains the effectiveness of language model fine-tuning. In Proc.
  59th Annual Meeting of the Association for Computational Linguistics
  7319--7328 (ACL, 2021).
\item
  Hu, E. J. et al. LoRA: low-rank adaptation of large language models.
  In Proc. 10th International Conference on Learning Representations
  https://openreview.net/forum?id=nZeVKeeFYf9 (2022).
\item
  Shaw, C. et al. Comparison of imputation strategies for incomplete
  longitudinal data in life-course epidemiology. Am. J. Epidemiol. 192,
  2075--2084 (2023).
\item
  Lazar, A., Jin, L., Spurlock, C. A., Wu, K., Sim, A. \& Todd, A.
  Evaluating the effects of missing values and mixed data types on
  social sequence clustering using t-SNE visualization. J. Data Inf.
  Qual. 11, 7:1--7:22 (2019).
\item
  Rindfuss, R. R. The young adult years: diversity, structural change,
  and fertility. Demography 28, 493--512 (1991).
\item
  Manning, W. D. Young adulthood relationships in an era of uncertainty:
  a case for cohabitation. Demography 57, 799--819 (2020).
\item
  Eliason, S. R., Mortimer, J. T. \& Vuolo, M. The transition to
  adulthood: life course structures and subjective perceptions. Soc.
  Psychol. Q. 78, 205--227 (2015).
\item
  Elzinga, C. H. Complexity of categorical time series. Sociol. Methods
  Res. 38, 463--481 (2010).
\item
  Van Winkle, Z. Family trajectories across time and space: increasing
  complexity in family life courses in Europe? Demography 55, 135--164
  (2018).
\item
  Jaccard, P. The distribution of the flora in the alpine zone. New
  Phytologist 11, 37--50 (1912).
\item
  Levenshtein, V. I. Binary codes capable of correcting deletions,
  insertions, and reversals. Sov. Phys. Dokl. 10, 707--710 (1966).
\item
  Vaserstein, L. N. Markov processes over denumerable products of
  spaces, describing large systems of automata. Probl. Peredachi Inf. 5,
  64--72 (1969).
\item
  Levandowsky, M. \& Winter, D. Distance between sets. Nature 234,
  34--35 (1971).
\item
  Begall, K. \& Grunow, D. Labour force transitions around first
  childbirth in the Netherlands. Eur. Sociol. Rev. 31, 697--712 (2015).
\item
  Hallqvist, J., Lynch, J., Bartley, M., Lang, T. \& Blane, D. Can we
  disentangle life course processes of accumulation, critical period and
  social mobility? An analysis of disadvantaged socio-economic positions
  and myocardial infarction in the Stockholm Heart Epidemiology Program.
  Soc. Sci. Med. 58, 1555--1562 (2004).
\item
  McCoy, R. T., Pavlick, E. \& Linzen, T. Right for the wrong reasons:
  diagnosing syntactic heuristics in natural language inference. Proc.
  57th Annu. Meet. Assoc. Comput. Linguist. 3428--3448 (2019).
\item
  Niven, T. \& Kao, H.-Y. Probing neural network comprehension of
  natural language arguments. Proc. 57th Annu. Meet. Assoc. Comput.
  Linguist. 4658--4664 (2019).
\item
  Brückner, H. \& Mayer, K. U. De-standardization of the life course:
  what it might mean? And if it means anything, whether it actually took
  place? Adv. Life Course Res. 9, 27--53 (2005).
\item
  Kendall, M. G. A new measure of rank correlation. Biometrika 30,
  81--93 (1938).
\item
  Hirschberg, D. S. A linear space algorithm for computing maximal
  common subsequences. Commun. ACM 18, 341--343 (1975).
\item
  Petroni, F. et al. Language models as knowledge bases? In Proceedings
  of the 2019 Conference on Empirical Methods in Natural Language
  Processing and the 9th International Joint Conference on Natural
  Language Processing (EMNLP-IJCNLP) 2463--2473 (Association for
  Computational Linguistics, 2019).
\item
  Wu, Z., Chen, Y., Kao, B. \& Liu, Q. Perturbed masking: parameter-free
  probing for analyzing and interpreting BERT. In Proceedings of the
  58th Annual Meeting of the Association for Computational Linguistics
  4166--4176 (Association for Computational Linguistics, 2020).
\item
  Talmor, A., Elazar, Y., Goldberg, Y. \& Berant, J. oLMpics -- on what
  language model pre-training captures. Trans. Assoc. Comput. Linguist.
  8, 743--758 (2020).
\item
  Zeiler, M. D. \& Fergus, R. Visualizing and understanding
  convolutional networks. In Computer Vision -- ECCV 2014 (eds Fleet, D.
  et al.) 818--833 (Springer, 2014).
\item
  Elder, G. H. Jr. Military times and turning points in
  men\textquotesingle s lives. Dev. Psychol. 22, 233--245 (1986).
\item
  Sampson, R. J. \& Laub, J. H. Socioeconomic achievement in the life
  course of disadvantaged men: military service as a turning point,
  circa 1940--1965. Am. Sociol. Rev. 61, 347--367 (1996).
\item
  Clark, A. E., Diener, E., Georgellis, Y. \& Lucas, R. E. Lags and
  leads in life satisfaction: a test of the baseline hypothesis. The
  Economic Journal 118, F222--F243 (2008).
\item
  Jacobson, L. S., LaLonde, R. J. \& Sullivan, D. G. Earnings losses of
  displaced workers. American Economic Review 83, 685--709 (1993).
\item
  Sullivan, D. \& von Wachter, T. Job displacement and mortality: an
  analysis using administrative data. The Quarterly Journal of Economics
  124, 1265--1306 (2009).
\item
  Stacherl, B. \& Sauzet, O. Chronic disease onset and wellbeing
  development: longitudinal analysis and the role of healthcare access.
  Eur. J. Public Health 34, 29--34 (2024).
\item
  Stöckel, J., van Exel, J. \& Brouwer, W. B. F. Adaptation in life
  satisfaction and self-assessed health to disability---evidence from
  the UK. Soc. Sci. Med. 328, 115996 (2023).
\item
  Gao, T., Fisch, A. \& Chen, D. Making pre-trained language models
  better few-shot learners. In Proceedings of the 59th Annual Meeting of
  the Association for Computational Linguistics and the 11th
  International Joint Conference on Natural Language Processing
  (ACL-IJCNLP) 3816--3830 (Association for Computational Linguistics,
  2021).
\item
  Roberts, A., Raffel, C. \& Shazeer, N. How much knowledge can you pack
  into the parameters of a language model? In Proceedings of the 2020
  Conference on Empirical Methods in Natural Language Processing (EMNLP)
  5418--5426 (Association for Computational Linguistics, 2020).
\item
  Hegselmann, S., Buendia, A., Lang, H., Agrawal, M., Jiang, X. \&
  Sontag, D. TabLLM: Few-shot Classification of Tabular Data with Large
  Language Models. Proc. 26th International Conference on Artificial
  Intelligence and Statistics (AISTATS) PMLR 206, 5549--5581 (2023).
\item
  Dinh, T., Zeng, Y., Zhang, R., Lin, Z., Gira, M., Rajput, S., Sohn,
  J., Papailiopoulos, D. \& Lee, K. LIFT: Language-Interfaced
  Fine-Tuning for Non-Language Machine Learning Tasks. Advances in
  Neural Information Processing Systems 35, 11763--11784 (2022).
\item
  Howard, J. \& Ruder, S. Universal Language Model Fine-tuning for Text
  Classification. Proc. 56th Annual Meeting of the Association for
  Computational Linguistics 1, 328--339 (2018).
\item
  Aisenbrey, S. \& Fasang, A. E. The interplay of work and family
  trajectories over the life course: Germany and the United States in
  comparison. Am. J. Sociol. 122, 1448--1484 (2017).
\item
  Bratberg, E., Davis, J., Mazumder, B., Nybom, M., Schnitzlein, D. D.
  \& Vaage, K. A comparison of intergenerational mobility curves in
  Germany, Norway, Sweden, and the US. Scand. J. Econ. 119, 72--101
  (2017).
\item
  Stovel, K. \& Bolan, M. Residential trajectories: using optimal
  alignment to reveal the structure of residential mobility. Sociol.
  Methods Res. 32, 559--598 (2004).
\item
  Massoglia, M. Incarceration as exposure: the prison, infectious
  disease, and other stress-related illnesses. J. Health Soc. Behav. 49,
  56--71 (2008).
\item
  Hinton, G., Vinyals, O. \& Dean, J. Distilling the knowledge in a
  neural network. Preprint at https://arxiv.org/abs/1503.02531 (2015).
\item
  Guo, C., Pleiss, G., Sun, Y. \& Weinberger, K. Q. On calibration of
  modern neural networks. In Proceedings of the 34th International
  Conference on Machine Learning 1321--1330 (PMLR, 2017).
\item
  Kuhn, L., Gal, Y. \& Farquhar, S. Semantic uncertainty: linguistic
  invariances for uncertainty estimation in natural language generation.
  In Proceedings of the 11th International Conference on Learning
  Representations (ICLR, 2023).
\item
  Romano, Y., Patterson, E. \& Candès, E. J. Conformalized quantile
  regression. In Advances in Neural Information Processing Systems 32,
  3538--3548 (NeurIPS, 2019).
\item
  Bolukbasi, T., Chang, K.-W., Zou, J. Y., Saligrama, V. \& Kalai, A. T.
  Man is to computer programmer as woman is to homemaker? Debiasing word
  embeddings. In Advances in Neural Information Processing Systems 29,
  4349--4357 (NeurIPS, 2016).
\item
  Caliskan, A., Bryson, J. J. \& Narayanan, A. Semantics derived
  automatically from language corpora contain human-like biases. Science
  356, 183--186 (2017).
\item
  Zhao, J., Wang, T., Yatskar, M., Ordonez, V. \& Chang, K.-W. Gender
  bias in coreference resolution: evaluation and debiasing methods. In
  Proceedings of the 2018 Conference of the North American Chapter of
  the Association for Computational Linguistics: Human Language
  Technologies Vol. 2, 15--20 (ACL, 2018).
\item
  Dettmers, T., Pagnoni, A., Holtzman, A. \& Zettlemoyer, L. QLoRA:
  Efficient Finetuning of Quantized LLMs. Adv. Neural Inf. Process.
  Syst. 36, 10088--10115 (2023).
\item
  Loshchilov, I. \& Hutter, F. Decoupled weight decay regularization. in
  International Conference on Learning Representations (ICLR, 2019).
\item
  Ma, J. \& Yarats, D. On the adequacy of untuned warmup for adaptive
  optimization. Proc. AAAI Conf. Artif. Intell. 35, 8828--8836 (2021).
\item
  Smith, S. L., Kindermans, P.-J., Ying, C. \& Le, Q. V.
  Don\textquotesingle t decay the learning rate, increase the batch
  size. in International Conference on Learning Representations (ICLR,
  2018).
\item
  Chen, T. \& Guestrin, C. XGBoost: a scalable tree boosting system. In
  Proceedings of the 22nd ACM SIGKDD International Conference on
  Knowledge Discovery and Data Mining 785--794 (ACM, 2016).
\item
  Hochreiter, S. \& Schmidhuber, J. Long short-term memory. Neural
  Comput. 9, 1735--1780 (1997).
\item
  Cho, K. et al. Learning phrase representations using RNN
  encoder--decoder for statistical machine translation. In Proceedings
  of the 2014 Conference on Empirical Methods in Natural Language
  Processing (EMNLP) 1724--1734 (Association for Computational
  Linguistics, 2014).
\item
  Yao, Y., Rosasco, L. \& Caponnetto, A. On early stopping in gradient
  descent learning. Constr. Approx. 26, 289--315 (2007).
\end{enumerate}
}

\par

%% file: supp_body.tex
\subsection*{Prompt template design}

The prompt followed a fixed five-section template:

\begin{Verbatim}
### System:
You are a Life Trajectory Specialist. You are trained to analyze
longitudinal life history data, including discrete events
(employment, family, health) and continuous state variables
(income, satisfaction), to perform complex forecasting and
reasoning tasks.

### Task Type:
{TASK_IDENTIFIER}

### Input Data:
{FORMATTED_INPUT}

### Instruction:
Subject: {gender}, Born: {birth_year}. {task_instruction}
Return the answer strictly as a valid JSON object.
Do not output any additional text, explanations, or markdown
formatting.

### Prediction:
{MODEL_OUTPUT}
\end{Verbatim}

The System section was identical across all tasks. The Task Type section
contained one of the 18 task identifiers (for example, NEXT\_EVENT,
ANOMALY\_DETECTION, REORDERING). The Instruction section embedded the
individual\textquotesingle s demographic information (gender and birth
year) and the task-specific directive. The Prediction section contained
the ground-truth JSON target during training, and was left empty at
inference time for the model to complete via greedy decoding.

\subsection*{Discrete event formatting}

Each life event was rendered as a single line of the form:

\begin{Verbatim}
- Event: {EVENT_TYPE} at age {age} ({year}). Details: {json_details}
\end{Verbatim}

The event-detail dictionary was serialized as JSON with keys sorted
alphabetically to ensure deterministic output. For example, a full-time
employment event might be rendered as:

\begin{Verbatim}
- Event: JOB_START at age 24 (1972). Details: {"employment_type":
  "Full-Time Employment", "occupation": "Mechanical Engineers",
  "position": "Skilled worker"}
\end{Verbatim}

Events without associated details (for example, DEATH or SEPARATION)
omitted the Details suffix. Masked events were rendered as
\texttt{- [MASKED EVENT]: Event hidden} for individually masked events,
or \texttt{- [MASKED EVENT]: Redacted Period (1985--1992)} for
contiguous temporal redactions.

\subsection*{State history formatting}

Continuous state measurements were grouped by calendar year and
consolidated into a single line per year. Within each line, state types
were sorted alphabetically and formatted as human-readable key-value
pairs. For example:

\begin{Verbatim}
- At age 35 (1983): Health Satisfaction: 7, Income Level: 42850.0,
  Job Prestige: 3: Limited autonomy of action, intermediate,
  Life Satisfaction: 8
\end{Verbatim}

Years without any valid state measurements were omitted. If no state
measurements were available for an individual up to the split point, the
section displayed No prior state measurements.

\subsection*{Input layout variants}

The Input Data section was dynamically structured according to the task
type. Most tasks used a single chronological event list under a Discrete
Event Sequence header, followed by a Continuous State History block. Two
task types required split inputs: CONNECT\_POINTS organized the input
into Known History (events up to the start point) and Target Future
State (the single anchor event), while IMPUTE\_RANGE organized the input
into History Before Gap and History After Gap. Tasks that did not use
state information (ANOMALY\_DETECTION, IMPLICIT\_REMOVAL, REORDERING,
STATE\_IMPUTATION) omitted the state-history block entirely.

\subsection*{Supplementary Figures}

\begin{figure}[tbp]
{\centering
\includegraphics[width=6.27014in,height=2.38333in]{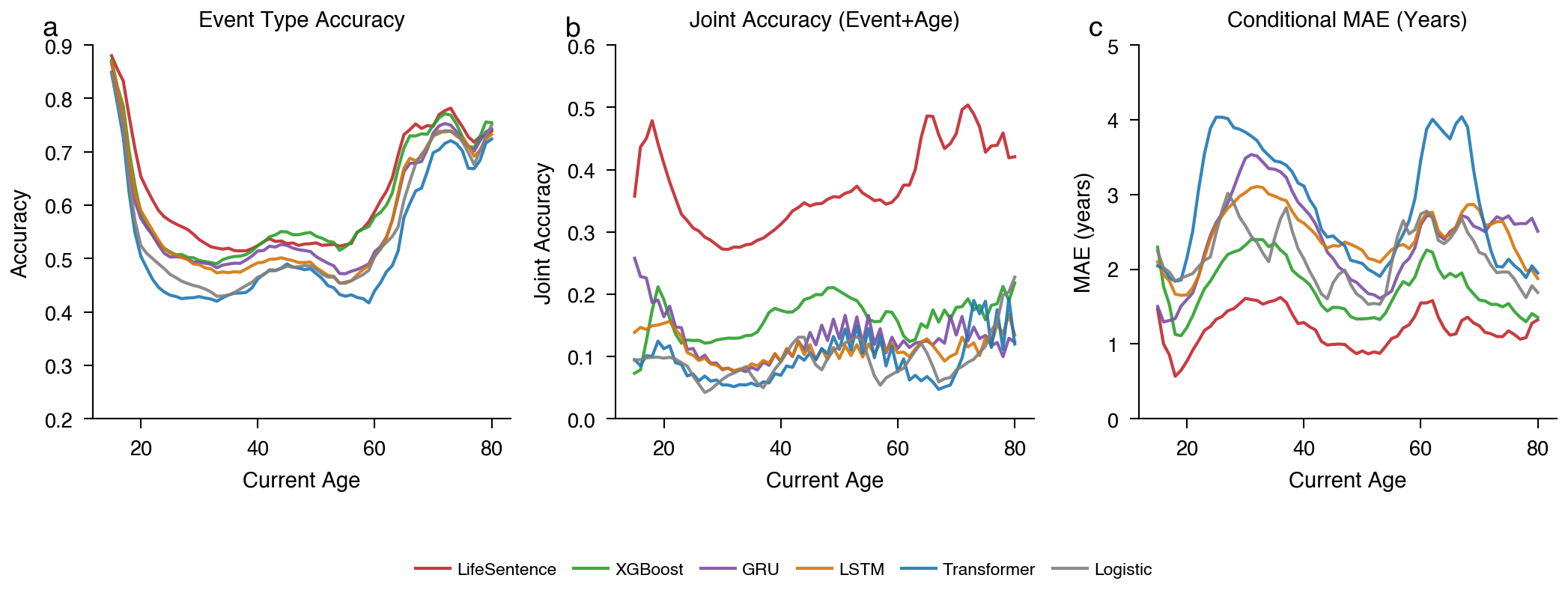}
\par}
\smallskip
{\small
\textbf{Supplementary Figure 1\,---\, Age-stratified predictive
performance on the next-event task.} (a) Event type accuracy, (b) joint
accuracy (correct event type and age within one year), and (c)
conditional mean absolute error (MAE, in years, computed only for
correctly predicted event types) as a function of the current age at
which the individual\textquotesingle s history is truncated. All models
exhibit reduced accuracy during the 20--30 age range, consistent with
this period\textquotesingle s characterization as a demographically
dense interval of concurrent life transitions. LifeSentence (blue)
maintains substantially higher performance across all age ranges and
metrics. The advantage is most pronounced for joint accuracy, where
LifeSentence sustains values above approximately 25\% even at peak
volatility (ages 20--30), while all baselines remain below 15\%.
Conditional MAE remains near or below one year for LifeSentence across
the full age range, whereas baseline models exhibit errors of two to
four years.
\par}
\end{figure}

\begin{figure}[tbp]
{\centering
\includegraphics[width=6.27014in,height=5.84444in]{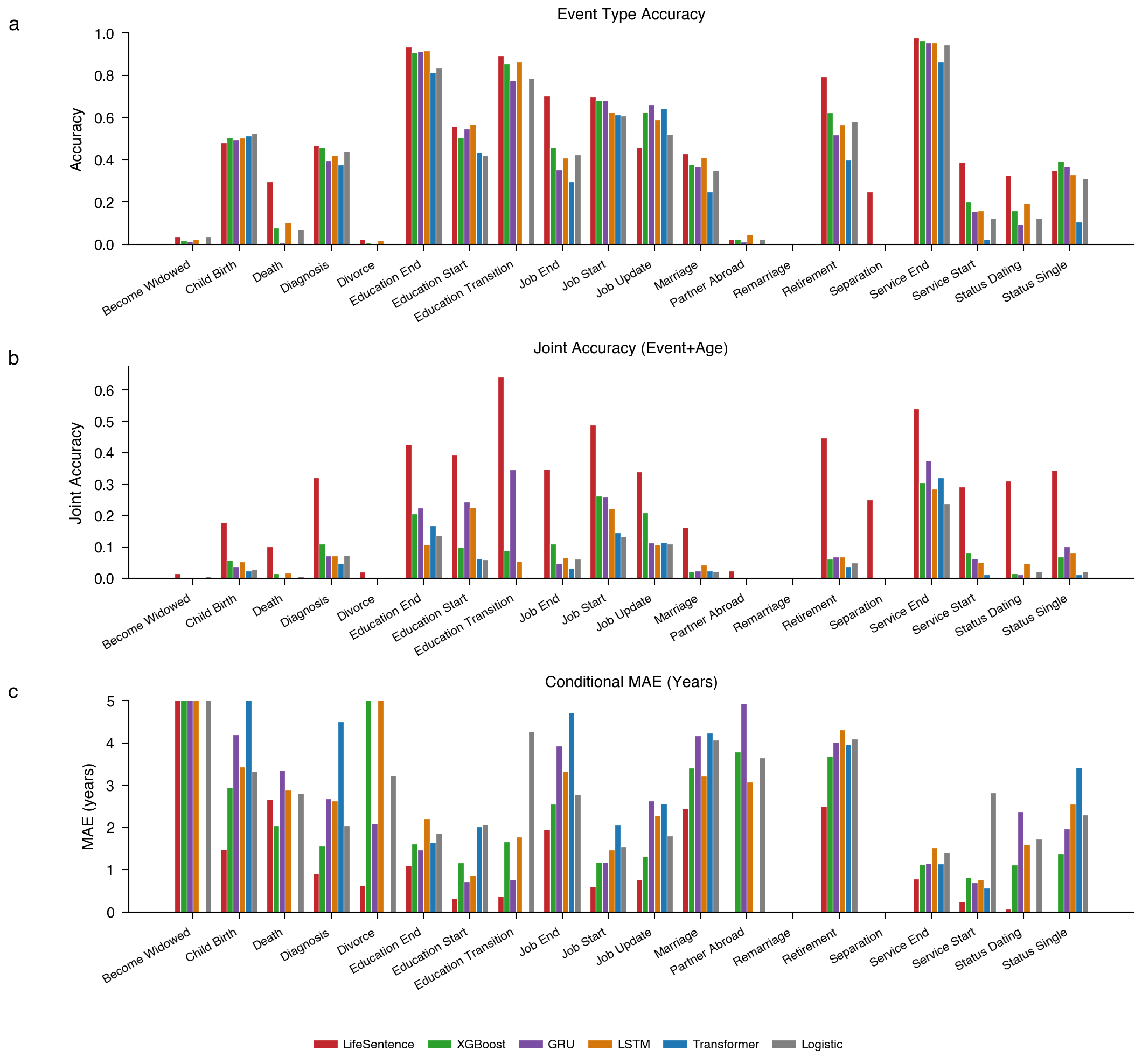}
\par}
\smallskip
{\small
\textbf{Supplementary Figure 2\,---\, Next-event prediction
performance stratified by event type and birth cohort}. (a--c)
Performance by event type. (a) Event type accuracy, (b) joint accuracy
(correct event type and age within one year), and (c) conditional mean
absolute error (MAE, in years) across all 15 event categories.
LifeSentence (blue) achieves the highest accuracy for the majority of
event types, with its advantage most pronounced for socially normative
events with well-defined temporal windows. Notably, divorce and death
remain difficult for all models, reflecting low base rates and high
temporal variance. LifeSentence\textquotesingle s conditional MAE is
consistently the lowest across event types, with near-zero errors for
service start and service end, indicating precise temporal placement for
institutionally bounded events. All models struggle with the timing of
relationship status changes (dating, single), consistent with the high
entropy of these transitions.
\par}
\end{figure}

\begin{figure}[tbp]
{\centering
\includegraphics[width=6.27014in,height=5.84444in]{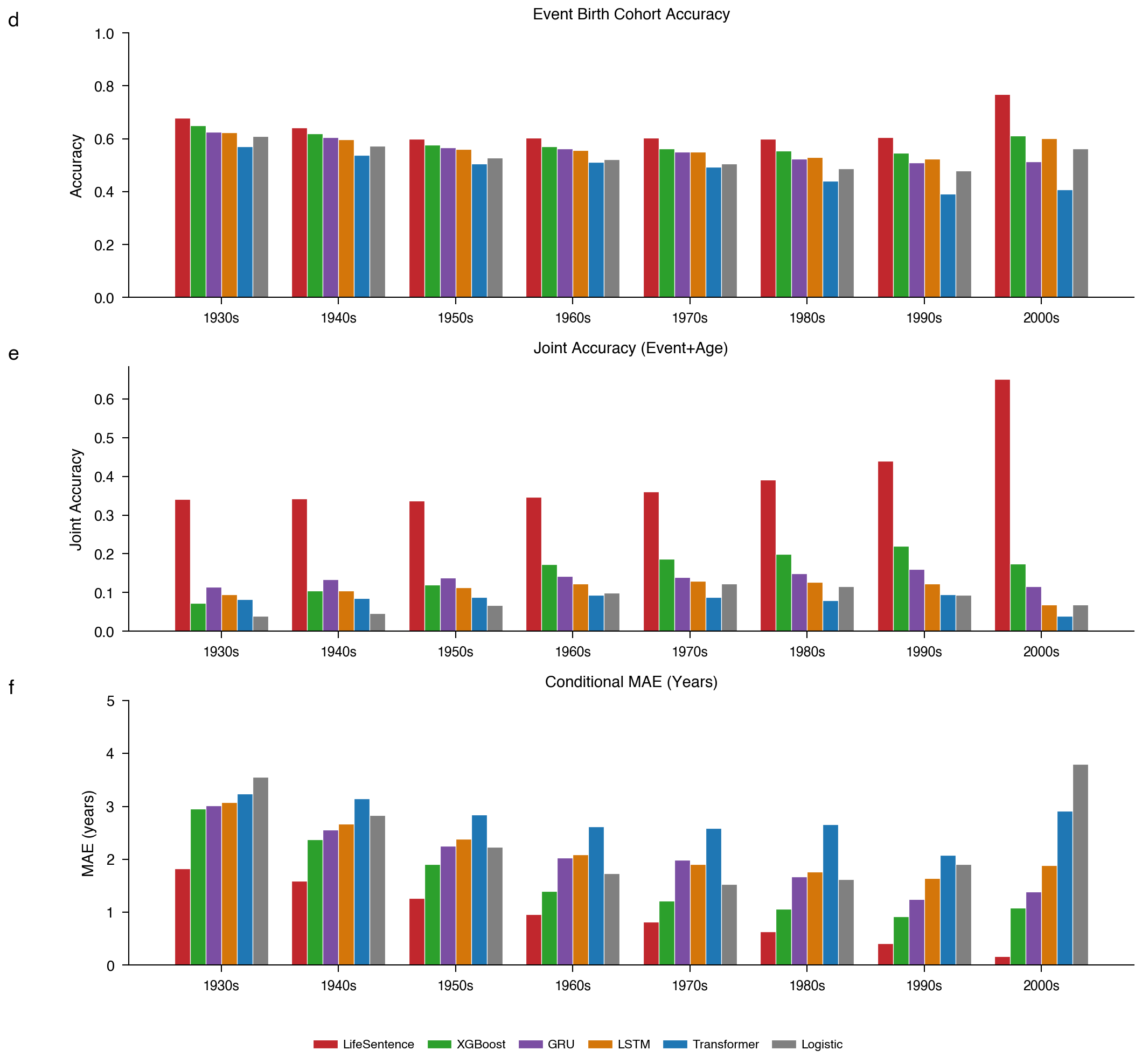}
\par}
\smallskip
{\small
\textbf{Supplementary Figure 2 (continued)\,---\, Next-event
prediction performance stratified by birth cohort} (d--f) Performance by
birth cohort. (d) Event type accuracy, (e) joint accuracy, and (f)
conditional MAE for individuals grouped by birth decade (1930s--2000s).
LifeSentence maintains a stable advantage across all cohorts, and its
joint accuracy and conditional MAE improve for more recent cohorts
(1980s--2000s), likely reflecting denser observation windows and more
complete event histories for younger individuals within the SOEP panel.
Baseline models show comparatively flat or inconsistent trends across
cohorts. The consistency of LifeSentence\textquotesingle s advantage
across both cohorts and event types indicates that the model captures
generalizable temporal dependencies rather than overfitting to
cohort-specific or event-specific patterns.
\par}
\end{figure}

\begin{figure}[tbp]
{\centering
\includegraphics[width=6.27014in,height=2.3in,keepaspectratio]{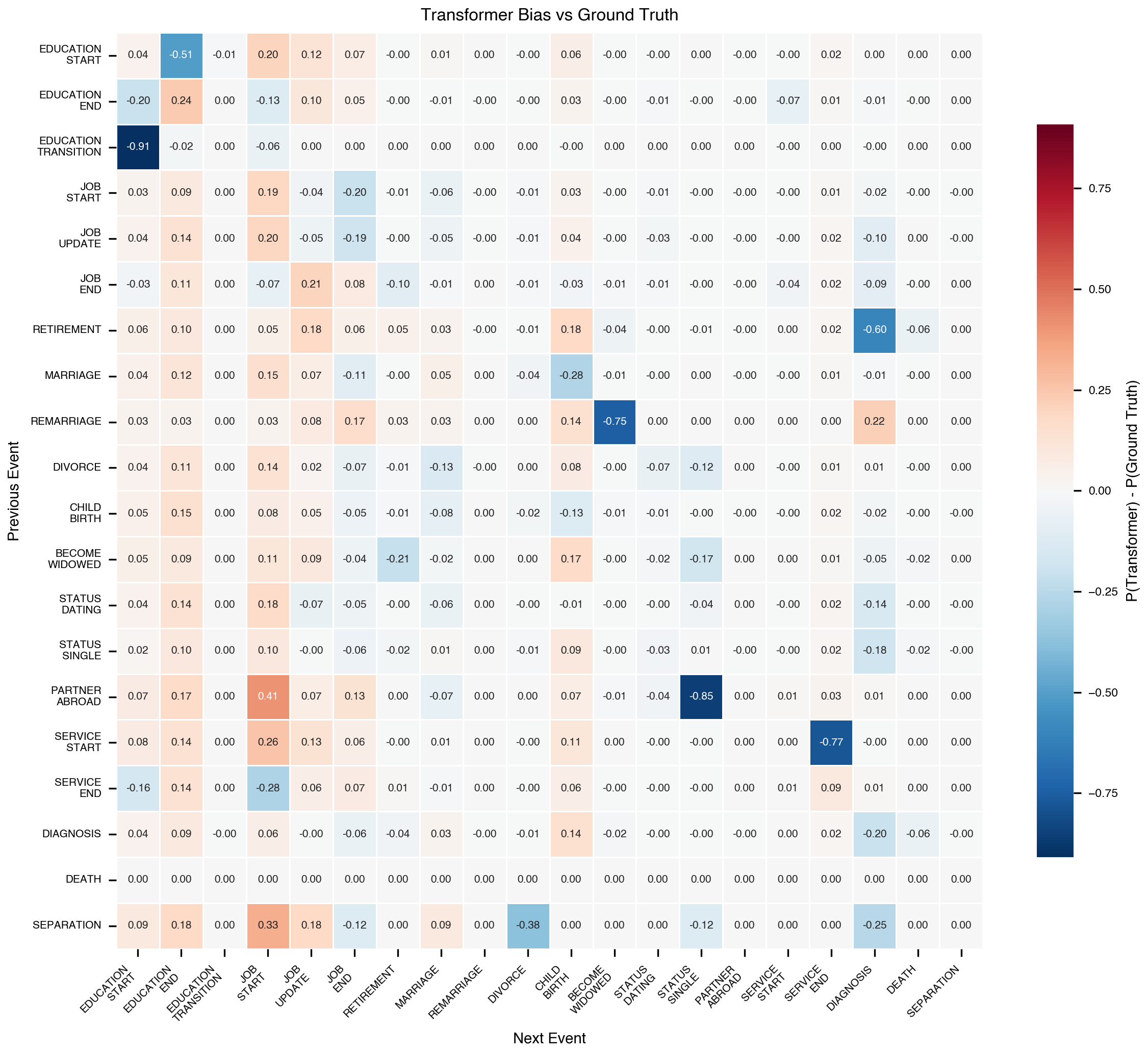}

\includegraphics[width=6.27014in,height=2.3in,keepaspectratio]{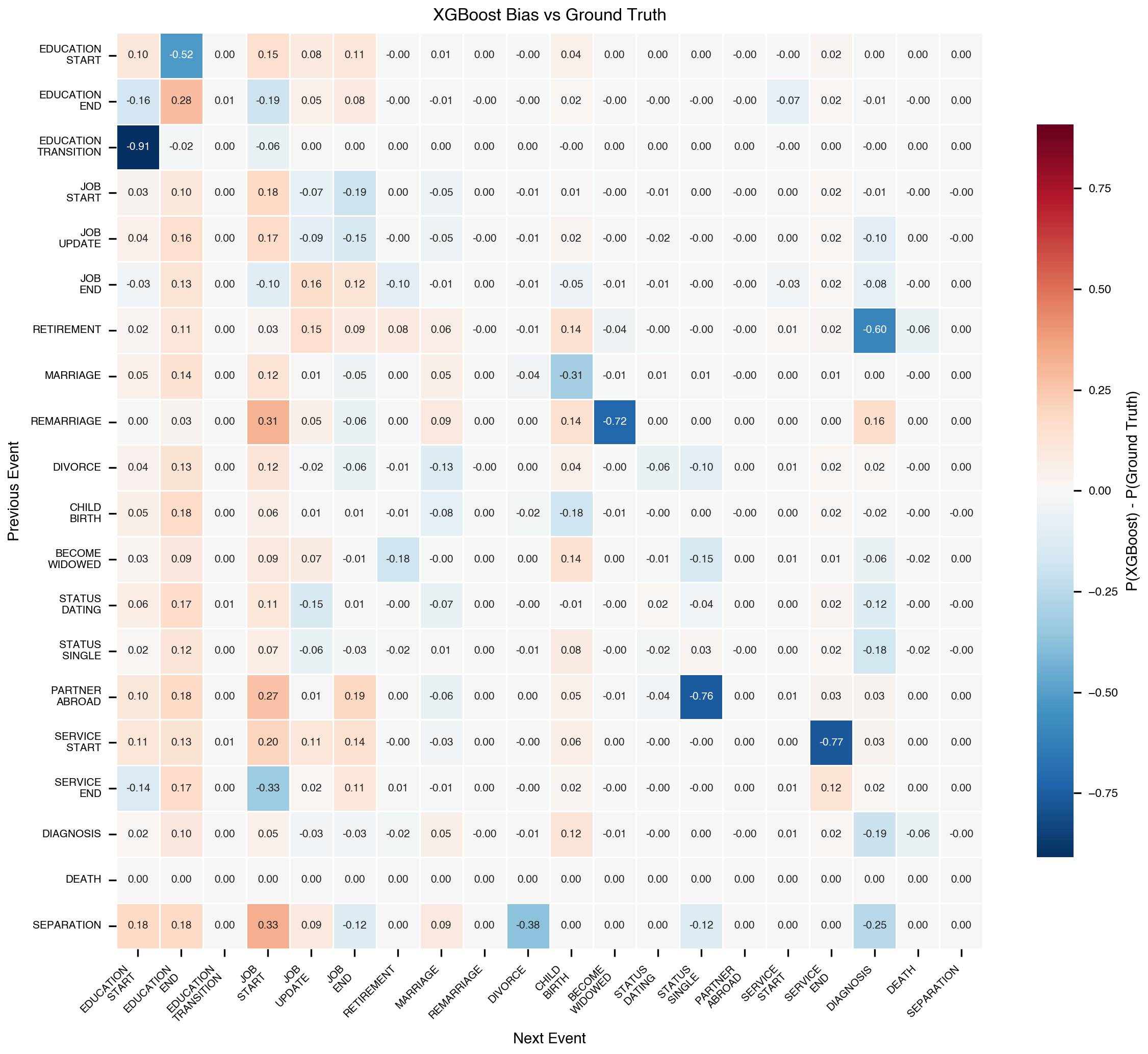}

\includegraphics[width=6.27014in,height=2.3in,keepaspectratio]{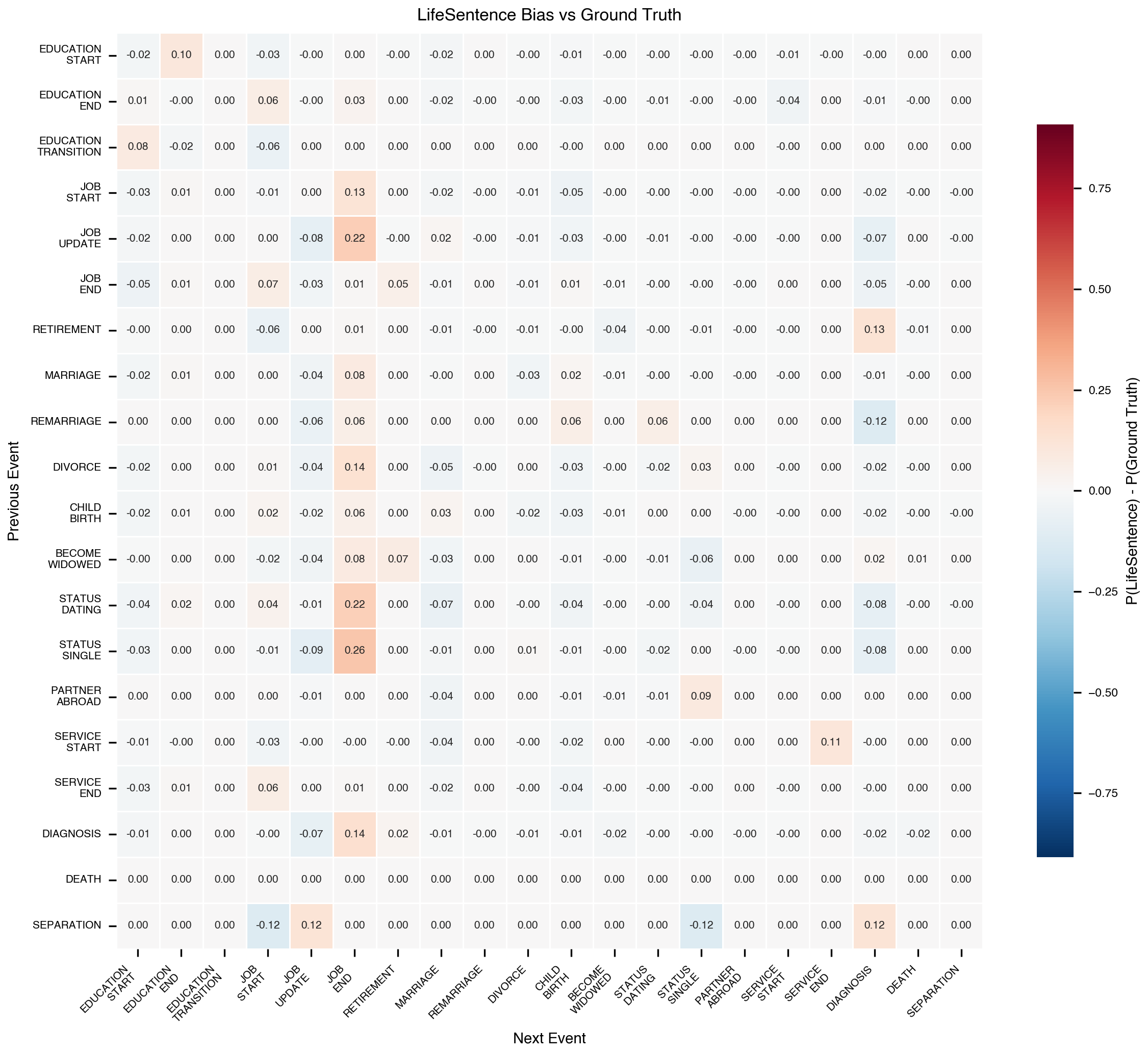}
\par}
\smallskip
{\small
\textbf{Supplementary Figure 3\,---\, Transition matrix residuals
for baseline and LifeSentence models.} Each panel displays the
difference between a model\textquotesingle s predicted transition
probabilities and the ground truth transition probabilities (P(Model) -
P(Ground Truth)) for all pairwise event sequences. Rows indicate the
preceding event, and columns indicate the next event. Red cells denote
overprediction (the model assigns higher probability to that transition
than observed empirically) and blue cells denote underprediction. All
three panels share a common colour scale (-.8 to +.8) to enable direct
visual comparison. (a/b) The residual matrices exhibit large errors
concentrated in deterministic transition pairs. The models massively
underpredict the SERVICE\_START $\rightarrow$ SERVICE\_END transition and the
RETIREMENT $\rightarrow$ DIAGNOSIS transition, indicating a failure to learn that
institutionally and biographically paired events follow each other with
high probability. (c) LifeSentence. The residual matrix is markedly
sparser and closer to zero. The largest remaining errors involve
moderate overprediction of JOB\_END following relationship status events
such as STATUS\_DATING (+0.22) and STATUS\_SINGLE (+0.26). Critically,
LifeSentence resolves the deterministic transition failures that
dominate the baseline models: the SERVICE\_START $\rightarrow$ SERVICE\_END error is
reduced to +0.11, confirming that the model has learned the closure
constraints of institutionally bounded event pairs.
\par}
\end{figure}

\begin{figure}[tbp]
{\centering
\includegraphics[width=6.27014in,height=2.77292in]{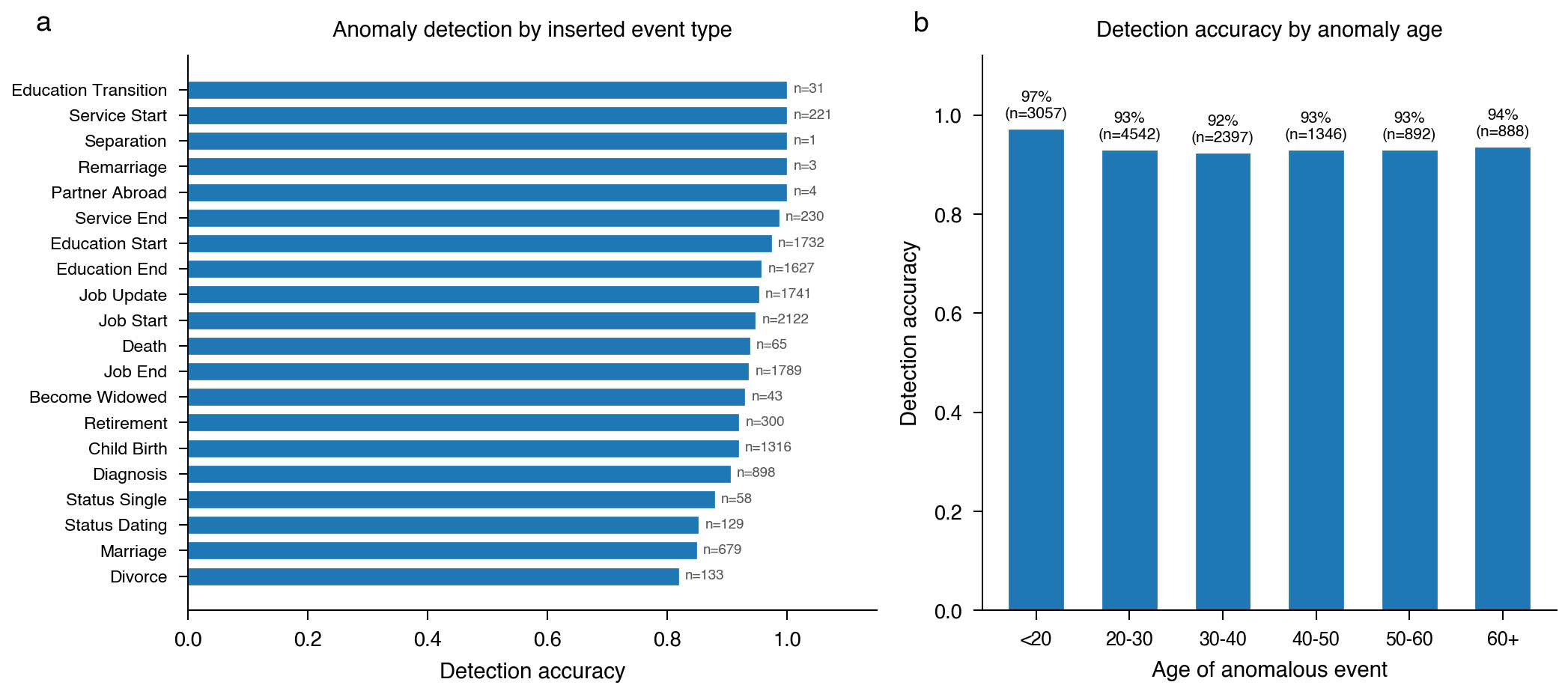}
\par}
\smallskip
{\small
\textbf{Supplementary Figure 4\,---\, Anomaly detection performance
by event type and age}. (a) Detection accuracy stratified by the type of
anomalous event inserted into the trajectory. Events embedded in
institutional sequences (e.g., SERVICE\_START, EDUCATION\_TRANSITION,
REMARRIAGE) are detected at near-perfect rates, whereas relationship
transitions (e.g., DIVORCE, MARRIAGE) prove more difficult to identify,
reflecting greater inter-individual variability in their timing and
occurrence. (b) Detection accuracy stratified by the age at which the
anomalous event was inserted. Accuracy is highest for events inserted
before age 20 (97\%), dips modestly during the 20--30 window of high
demographic activity (92\%), and stabilises at approximately 93\% across
subsequent life stages. Sample sizes are shown in parentheses. Overall
accuracy: 93.93\%.
\par}
\end{figure}

\begin{figure}[tbp]
{\centering
\includegraphics[width=6.27014in,height=4.47153in]{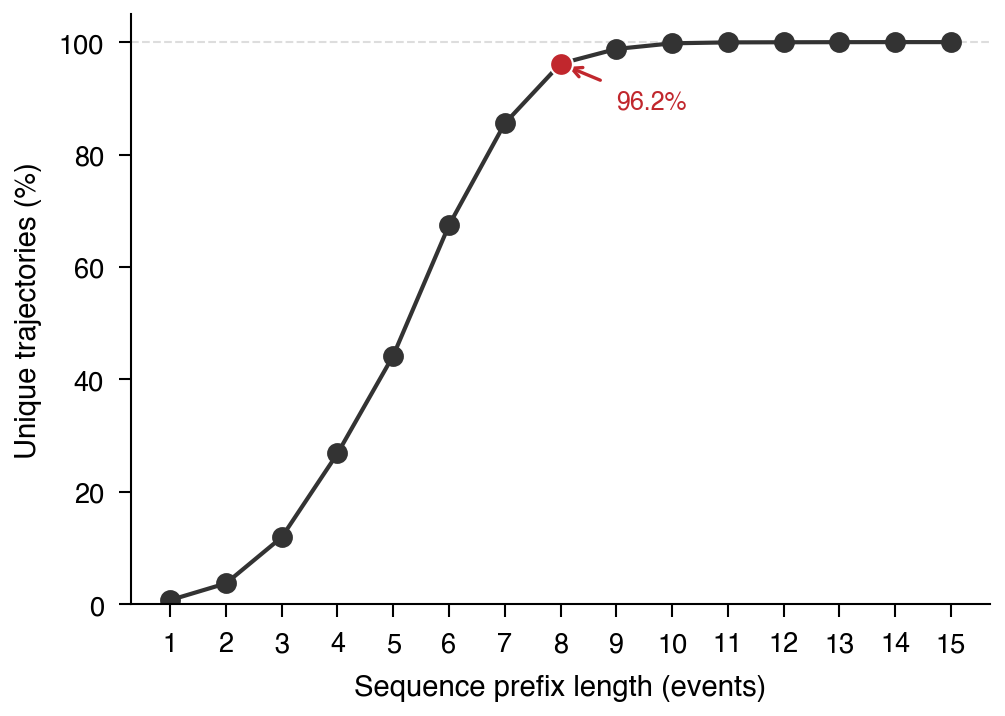}
\par}
\smallskip
{\small
\textbf{Supplementary Figure 5\,---\, Combinatorial complexity of
human life trajectories}. Percentage of unique event-type sequences in
the dataset as a function of sequence prefix length (number of events).
Uniqueness rises steeply: by the 8th event, 96.2\% of individuals
possess a life trajectory shared by no other person in the dataset, and
by the 10th event the figure approaches 99.8\%. This combinatorial
growth means that the model cannot rely on memorised templates when
generating complete trajectories. Each predicted biography must be
constructed from learned principles of temporal and causal structure
rather than retrieved from training examples, underscoring the
difficulty of the trajectory generation task.
\par}
\end{figure}

\begin{table}[tbp]
{\centering
\begin{adjustbox}{max width=\textwidth}
\begin{tabular}{@{}
  >{\raggedright\arraybackslash}p{(\linewidth - 8\tabcolsep) * \real{0.1491}}
  >{\raggedright\arraybackslash}p{(\linewidth - 8\tabcolsep) * \real{0.2096}}
  >{\raggedright\arraybackslash}p{(\linewidth - 8\tabcolsep) * \real{0.1996}}
  >{\raggedright\arraybackslash}p{(\linewidth - 8\tabcolsep) * \real{0.2395}}
  >{\raggedright\arraybackslash}p{(\linewidth - 8\tabcolsep) * \real{0.2022}}@{}}
\toprule
\textbf{Model} & \textbf{Jaccard Similarity} & \textbf{Multiset Jaccard}
& \textbf{Levenshtein Distance} & \textbf{Wasserstein Distance} \\
\midrule
\textbf{LifeSentence} & 0.627 {[}0.625-0.629{]} & 0.437
{[}0.435-0.439{]} & 8.103 {[}8.032-8.174{]} & 5.211 {[}5.168-5.252{]} \\
Transformer & 0.369 {[}0.367-0.372{]} & 0.092 {[}0.091-0.093{]} & 27.225
{[}27.202-27.247{]} & 11.580 {[}11.511-11.649{]} \\
LSTM & 0.528 {[}0.525-0.530{]} & 0.143 {[}0.142-0.144{]} & 26.189
{[}26.163-26.216{]} & 19.402 {[}19.279-19.521{]} \\
GRU & 0.524 {[}0.522-0.526{]} & 0.233 {[}0.232-0.234{]} & 19.456
{[}19.383-19.528{]} & 14.674 {[}14.602-14.750{]} \\
XGBoost & 0.520 {[}0.518-0.522{]} & 0.275 {[}0.273-0.277{]} & 19.867
{[}19.716-20.019{]} & 7.490 {[}7.438-7.541{]} \\
Logistic & 0.496 {[}0.494-0.498{]} & 0.174 {[}0.172-0.175{]} & 32.315
{[}32.162-32.479{]} & 12.756 {[}12.674-12.838{]} \\
\bottomrule
\end{tabular}
\end{adjustbox}
\par}
\smallskip
{\small
\textbf{Supplementary Table 2\,---\, Structural fidelity of
generated life trajectories}. Comparison of LifeSentence against
baseline models on the entire trajectory generation task. Metrics
include Jaccard Similarity (measuring the overlap of unique event
types), Multiset Jaccard (accounting for frequency of repeated events),
Levenshtein Distance (minimum edit distance to transform the predicted
sequence into the ground truth), and Wasserstein Distance (measuring the
cost to move predicted events to their correct temporal positions). 95\%
confidence intervals (in brackets) were calculated via 1000 bootstrapped
samples. LifeSentence outperforms all baselines across all metrics,
indicating superior narrative and temporal coherence.
\par}
\end{table}

\begin{figure}[tbp]
{\centering
\includegraphics[width=6.27014in,height=4.48056in]{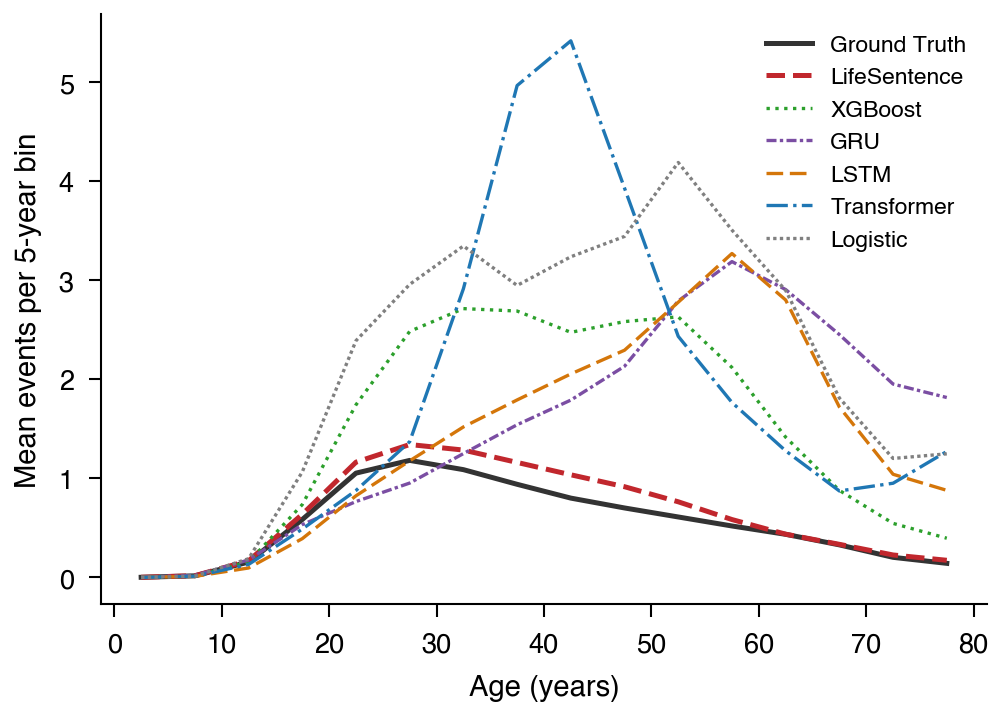}
\par}
\smallskip
{\small
\textbf{Supplementary Figure 6\,---\, Temporal density of life
events.} Comparison of the mean number of predicted events per year
across age bins for LifeSentence, Ground Truth, and baseline models.
LifeSentence accurately reproduces the non-linear peak in demographic
activity between ages 25 and 30, matching the ground truth distribution.
\par}
\end{figure}

\begin{figure}[tbp]
{\centering
\includegraphics[width=6.27014in,height=2.14375in]{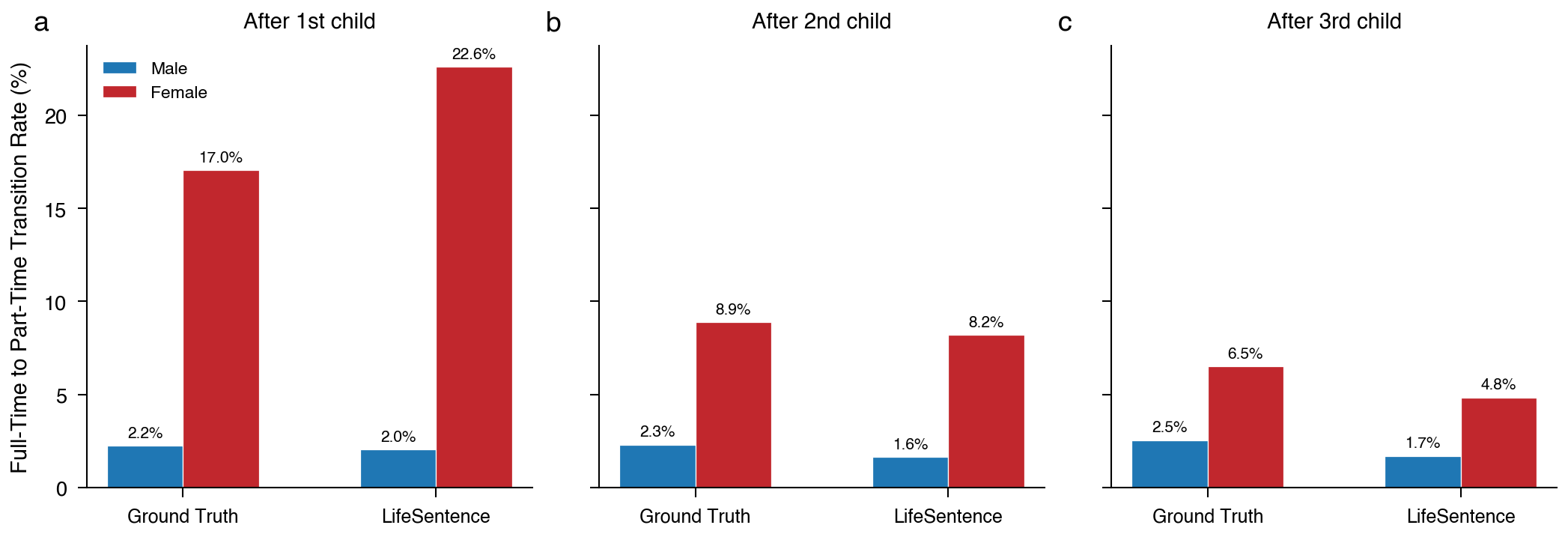}
\par}
\smallskip
{\small
\textbf{Supplementary Figure 7} \textbf{\textbar{} Part-Time Work
Transition After Childbirth Across Gender.} The proportion of
individuals who transition from full-time to part-time employment
specifically following the birth of child, stratified by sex. The model
correctly predicts a significantly higher transition rate for females
compared to males.
\par}
\end{figure}

\begin{figure}[tbp]
{\centering
\includegraphics[width=6.27014in,height=3.04236in]{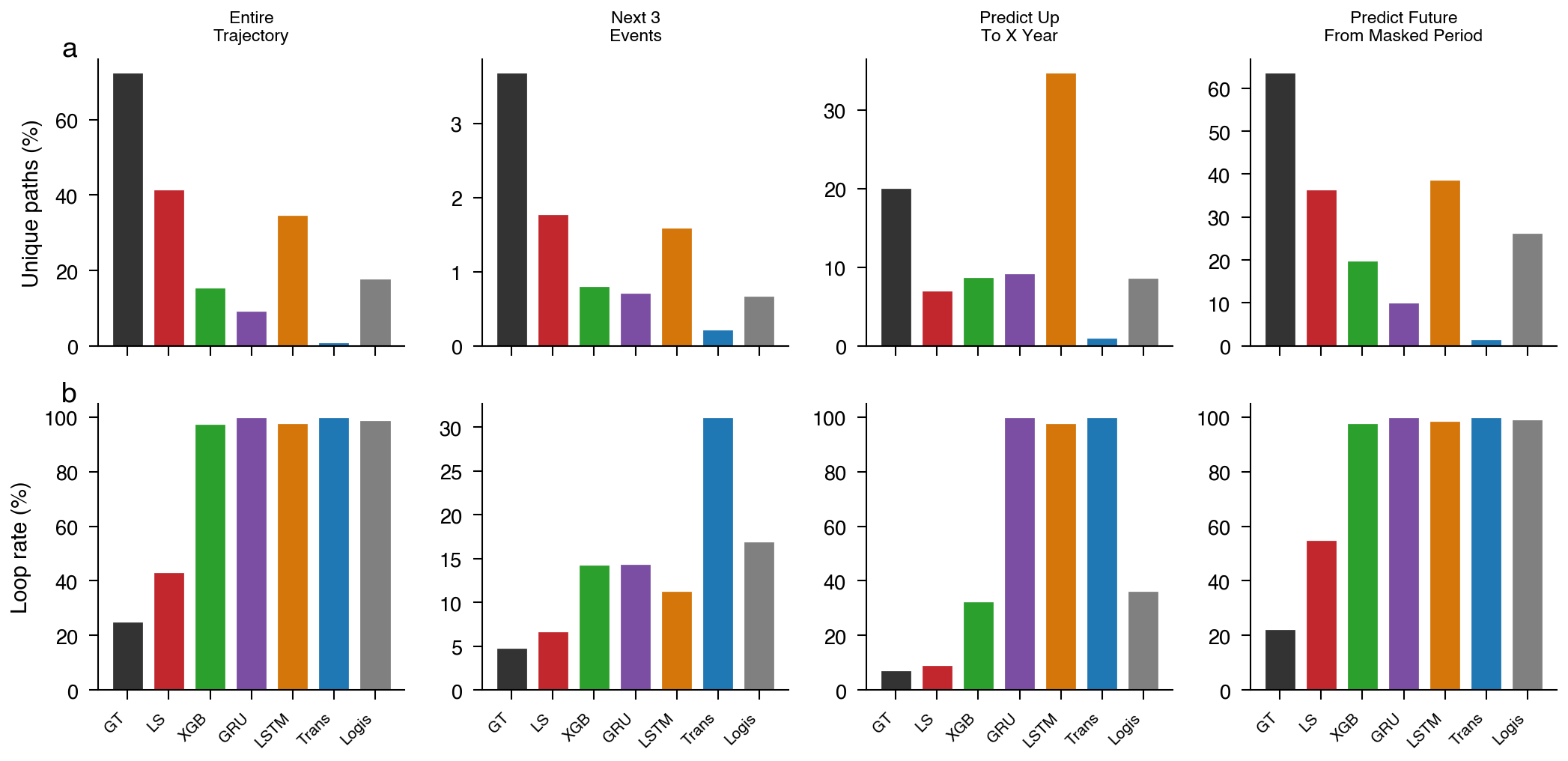}
\par}
\smallskip
{\small
\textbf{Supplementary Figure 8\,---\,} \textbf{Comparative analysis
of generative diversity and structural integrity}. (a) Comparison of
unique life path diversity, showing that LifeSentence generates a wider
array of distinct biographical sequences compared to the Transformer
baseline, which suffers from mode collapse. (b) Frequency of repetitive
loop artifacts (e.g., illogical repetitions of the same life event or
patterns). LifeSentence demonstrates a marked reduction in looping
behavior.
\par}
\end{figure}

\begin{table}[tbp]
{\centering
\begin{adjustbox}{max width=\textwidth}
\begin{tabular}{@{}
  >{\raggedright\arraybackslash}p{(\linewidth - 8\tabcolsep) * \real{0.1691}}
  >{\raggedright\arraybackslash}p{(\linewidth - 8\tabcolsep) * \real{0.1696}}
  >{\raggedright\arraybackslash}p{(\linewidth - 8\tabcolsep) * \real{0.2195}}
  >{\raggedright\arraybackslash}p{(\linewidth - 8\tabcolsep) * \real{0.2195}}
  >{\raggedright\arraybackslash}p{(\linewidth - 8\tabcolsep) * \real{0.2222}}@{}}
\toprule
Task & Model & Event Accuracy & Joint Accuracy & Age MAE \\
\midrule
Next 3 Events & LifeSentence & 0.462 {[}0.459-0.465{]} & 0.200
{[}0.198-0.202{]} & 1.74 {[}1.71-1.77{]} \\
Next 3 Events & XGBoost & 0.437 {[}0.434-0.440{]} & 0.083
{[}0.081-0.084{]} & 2.57 {[}2.54-2.60{]} \\
Next 3 Events & GRU & 0.425 {[}0.422-0.428{]} & 0.078 {[}0.077-0.080{]}
& 3.24 {[}3.21-3.28{]} \\
Next 3 Events & LSTM & 0.419 {[}0.416-0.422{]} & 0.059 {[}0.058-0.061{]}
& 3.29 {[}3.26-3.32{]} \\
Next 3 Events & Transformer & 0.384 {[}0.381-0.387{]} & 0.054
{[}0.052-0.055{]} & 3.34 {[}3.31-3.37{]} \\
Next 3 Events & Logistic & 0.397 {[}0.394-0.399{]} & 0.052
{[}0.050-0.053{]} & 3.20 {[}3.16-3.23{]} \\
Next Specific Type & LifeSentence & --- & 0.213 {[}0.209-0.218{]} & 4.23
{[}4.18-4.29{]} \\
\bottomrule
\end{tabular}
\end{adjustbox}
\par}
\smallskip
{\small
\textbf{Supplementary Table 3a\,---\, Short-horizon and
event-targeted constrained prediction.}

Next-3-Events task: accuracy of predicting the next three life events in
sequence following a random truncation point in the
individual\textquotesingle s history. Event Accuracy measures whether
each of the three predicted event types matches the ground truth event
at the same position. Joint Accuracy measures the probability that a
predicted event is both the correct type and occurs at the correct age
(within rounding to nearest year). Age MAE reports the mean absolute age
error conditional on correct event type prediction. Next Specific Type
task (LifeSentence only): the model is given the target event type and
asked to predict only when it will occur; event accuracy is therefore
100\% by construction and not reported. Joint Accuracy for this task is
the proportion of predictions within one year of the ground truth age.
LifeSentence outperforms all baselines on all Next-3-Events metrics. The
Next Specific Type task is evaluable only for LifeSentence as it
requires natural-language instruction following to specify the target
event type. 95\% confidence intervals (in brackets) were calculated via
1,000 bootstrap samples across all three prediction positions.
\par}
\end{table}

\begin{table}[tbp]
{\centering
\begin{adjustbox}{max width=\textwidth}
\begin{tabular}{@{}
  >{\raggedright\arraybackslash}p{(\linewidth - 10\tabcolsep) * \real{0.1655}}
  >{\raggedright\arraybackslash}p{(\linewidth - 10\tabcolsep) * \real{0.1405}}
  >{\raggedright\arraybackslash}p{(\linewidth - 10\tabcolsep) * \real{0.1964}}
  >{\raggedright\arraybackslash}p{(\linewidth - 10\tabcolsep) * \real{0.1568}}
  >{\raggedright\arraybackslash}p{(\linewidth - 10\tabcolsep) * \real{0.1814}}
  >{\raggedright\arraybackslash}p{(\linewidth - 10\tabcolsep) * \real{0.1594}}@{}}
\toprule
\begin{minipage}[b]{\linewidth}\centering
\textbf{Task}
\end{minipage} & \begin{minipage}[b]{\linewidth}\centering
\textbf{Model}
\end{minipage} & \begin{minipage}[b]{\linewidth}\centering
\textbf{Constraint Satisfaction}
\end{minipage} & \begin{minipage}[b]{\linewidth}\centering
\textbf{Jaccard}
\end{minipage} & \begin{minipage}[b]{\linewidth}\centering
\textbf{Levenshtein}
\end{minipage} & \begin{minipage}[b]{\linewidth}\centering
\textbf{Wasserstein}
\end{minipage} \\
\midrule
Predict Until Event & LifeSentence & 0.993 {[}0.992--0.994{]} & 0.650
{[}0.646--0.653{]} & 4.19 {[}4.11--4.26{]} & 4.67 {[}4.60--4.73{]} \\
Predict Until Event & LSTM & 0.792 {[}0.786--0.797{]} & 0.543
{[}0.540--0.547{]} & 9.31 {[}9.19--9.43{]} & 7.59 {[}7.50--7.67{]} \\
Predict Until Event & XGBoost & 0.754 {[}0.748--0.760{]} & 0.542
{[}0.538--0.545{]} & 8.78 {[}8.66--8.90{]} & 6.18 {[}6.12--6.25{]} \\
Predict Until Event & GRU & 0.713 {[}0.706--0.719{]} & 0.527
{[}0.524--0.531{]} & 11.88 {[}11.73--12.04{]} & 7.43 {[}7.34--7.52{]} \\
Predict Until Event & Logistic & 0.600 {[}0.594--0.607{]} & 0.495
{[}0.492--0.498{]} & 13.40 {[}13.25--13.55{]} & 9.14 {[}9.04--9.25{]} \\
Predict Until Event & Transformer & 0.446 {[}0.440--0.452{]} & 0.398
{[}0.394--0.401{]} & 17.37 {[}17.20--17.53{]} & 7.94 {[}7.86--8.02{]} \\
& & & & & \\
Predict Up To Year & LifeSentence & 0.917 {[}0.915--0.920{]} & 0.559
{[}0.556--0.562{]} & 2.25 {[}2.24--2.27{]} & 1.84 {[}1.81--1.87{]} \\
Predict Up To Year & XGBoost & 0.443 {[}0.438--0.448{]} & 0.441
{[}0.438--0.444{]} & 5.24 {[}5.17--5.30{]} & 2.14 {[}2.11--2.16{]} \\
Predict Up To Year & GRU & 0.024 {[}0.022--0.025{]} & 0.402
{[}0.399--0.404{]} & 4.71 {[}4.66--4.76{]} & 2.66 {[}2.63--2.69{]} \\
Predict Up To Year & Logistic & 0.358 {[}0.353--0.363{]} & 0.382
{[}0.379--0.384{]} & 5.66 {[}5.60--5.73{]} & 2.39 {[}2.36--2.41{]} \\
Predict Up To Year & Transformer & 0.096 {[}0.093--0.099{]} & 0.311
{[}0.308--0.314{]} & 7.48 {[}7.41--7.55{]} & 3.12 {[}3.07--3.16{]} \\
Predict Up To Year & LSTM & 0.038 {[}0.036--0.040{]} & 0.377
{[}0.374--0.379{]} & 4.25 {[}4.21--4.29{]} & 2.60 {[}2.57--2.63{]} \\
& & & & & \\
Snapshot Year & LifeSentence & 99.9\% {[}99.9--100.0\%{]} & 0.429
{[}0.425--0.434{]} & 0.84 {[}0.84--0.85{]} & \\
Connect Points & LifeSentence & 94.6\% {[}93.9--95.3\%{]} & 0.587
{[}0.580--0.595{]} & 7.08 {[}6.85--7.32{]} & 7.91 {[}7.75--8.07{]} \\
\bottomrule
\end{tabular}
\end{adjustbox}
\par}
\smallskip
{\small
\textbf{Supplementary Table 3b\,---\, Sequence generation under
temporal and semantic constraints.} Predict Until Event: the model
generates all future life events up to and including a specified
milestone event (death, retirement, or marriage). Constraint
Satisfaction reports whether the trajectory generates up to the
appropriate event/year. Predict Up To Year: the model generates all
events occurring before a specified calendar year. Jaccard measures
overlap of unique predicted event types against ground truth;
Levenshtein Distance measures the minimum sequence edit distance on
event-type strings; Wasserstein Distance measures the temporal
displacement cost of optimally matching predicted event ages to
ground-truth positions. LifeSentence achieves the highest Jaccard and
lowest Levenshtein and Wasserstein distances across all tasks where
comparison is possible. Baseline models satisfy the Predict Until Event
milestone constraint at rates between 44.6\% (Transformer) and 79.2\%
(LSTM) --- substantially above zero, but achieved through overgeneration
rather than targeted termination, as reflected in their substantially
worse Jaccard and Levenshtein scores. Snapshot Year and Connect Points
are evaluable only for LifeSentence, as both tasks require semantic
instruction following: Snapshot Year asks the model to generate only
events occurring during a specified future calendar year, while Connect
Points provides an early life history and a single distant anchor event
and asks the model to generate the intervening multi-decade trajectory.
95\% confidence intervals (in brackets) were calculated via 1,000
bootstrap samples.
\par}
\end{table}

\begin{figure}[tbp]
{\centering
\includegraphics[width=6.27014in,height=5.02778in]{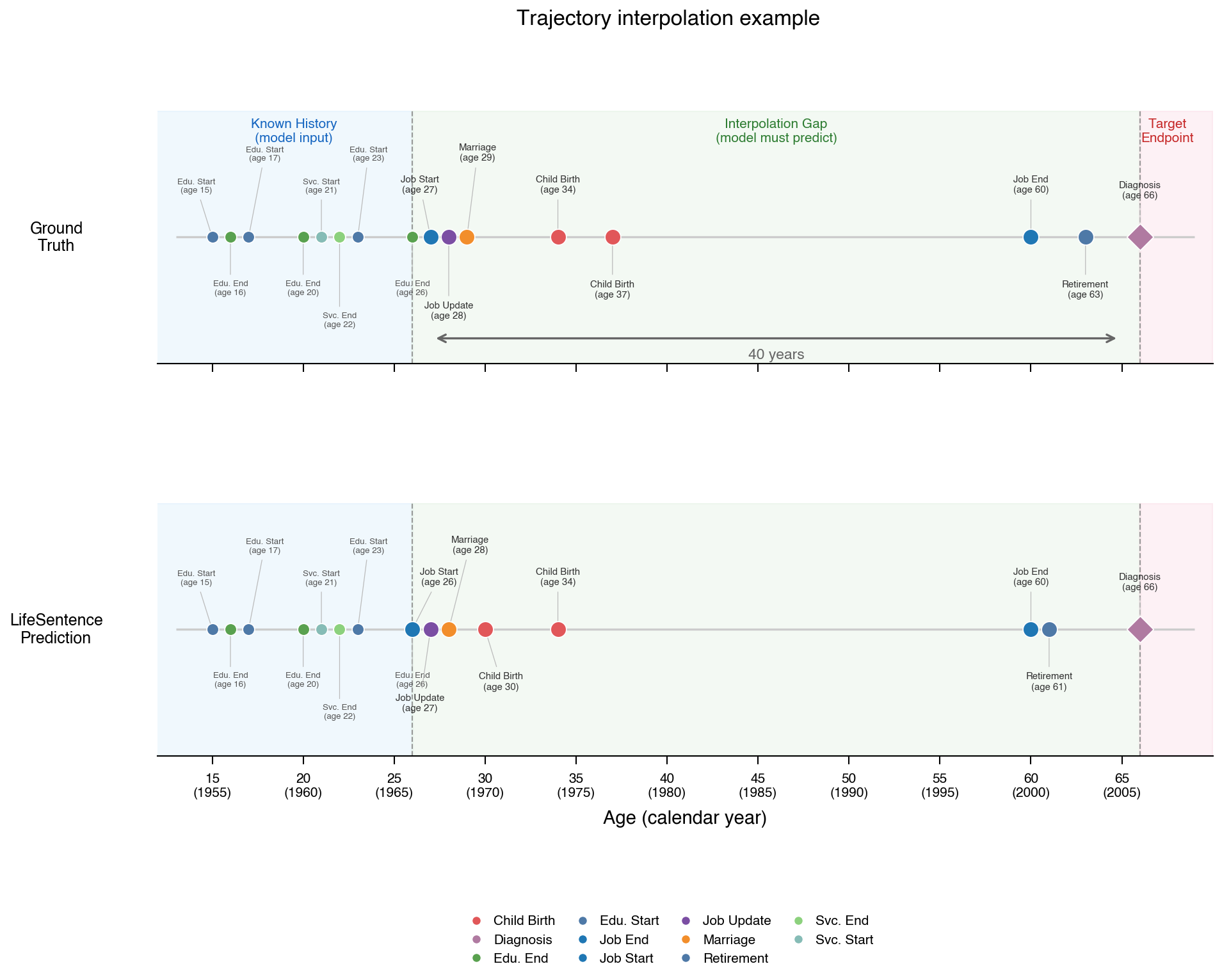}
\par}
\smallskip
{\small
\textbf{Supplementary Figure 9\,---\, Trajectory interpolation
across a 40-year observational gap.} Prediction for a single individual
(male, born 1940), in which the model receives the observed history up
to age 25 (left of dashed line) and a single target endpoint (Diagnosis
at age 66, red diamond) and must generate the intervening sequence of
life events spanning four decades. Top row: ground truth trajectory.
Bottom row: LifeSentence prediction. The model recovers every event type
in the correct order (Levenshtein distance = 0, Jaccard = 1.00),
reconstructing the canonical male life-course arc --- job entry and
career updating in the late 20s, marriage followed by two childbirths
through the 30s, a long period of occupational stability, and job end
preceding retirement in the early 60s --- with a mean temporal
displacement of only 1.71 years (Wasserstein distance). The largest
timing errors occur in the family formation window, where the model
places marriage and first childbirth approximately four years earlier
than observed (predicted ages 28 and 30 versus ground truth ages 29 and
34). This example illustrates that when provided with early biographical
context and a late-life anchor, LifeSentence can reconstruct the
intermediate trajectory with high fidelity, leveraging the structural
constraints of the life course --- that education precedes employment,
which precedes family formation, which precedes retirement --- to fill a
40-year gap.
\par}
\end{figure}

\begin{table}[tbp]
{\centering
\begin{adjustbox}{max width=\textwidth}
\begin{tabular}{@{}
  >{\raggedright\arraybackslash}p{(\linewidth - 6\tabcolsep) * \real{0.2042}}
  >{\raggedright\arraybackslash}p{(\linewidth - 6\tabcolsep) * \real{0.2649}}
  >{\raggedright\arraybackslash}p{(\linewidth - 6\tabcolsep) * \real{0.2450}}
  >{\raggedright\arraybackslash}p{(\linewidth - 6\tabcolsep) * \real{0.2859}}@{}}
\toprule
Model & \textbf{Masked History} & \textbf{Implicit Gap} & \textbf{Masked
Period} \\
\midrule
LifeSentence & 0.539 {[}0.534-0.544{]} & 0.496 {[}0.491-0.501{]} & 0.455
{[}0.448-0.461{]} \\
XGBoost & 0.437 {[}0.432-0.443{]} & 0.433 {[}0.428-0.438{]} & 0.363
{[}0.357-0.369{]} \\
GRU & 0.424 {[}0.419-0.429{]} & 0.421 {[}0.416-0.427{]} & 0.352
{[}0.346-0.358{]} \\
LSTM & 0.416 {[}0.411-0.421{]} & 0.413 {[}0.407-0.418{]} & 0.338
{[}0.332-0.344{]} \\
Transformer & 0.382 {[}0.377-0.387{]} & 0.376 {[}0.371-0.381{]} & 0.324
{[}0.317-0.329{]} \\
Logistic & 0.396 {[}0.390-0.401{]} & 0.390 {[}0.385-0.395{]} & 0.324
{[}0.318-0.330{]} \\
\bottomrule
\end{tabular}
\end{adjustbox}
\par}
\smallskip
{\small
\textbf{Supplementary Table 4\,---\,} \textbf{Robustness of
next-event prediction under missing historical information.} Next-event
type accuracy for LifeSentence and baseline models across three masking
tasks of increasing severity. Masked History: random individual events
in the history are replaced with a {[}MASKED EVENT{]} token, preserving
sequence length but removing content. Implicit Gap: events are silently
deleted without markers, simulating realistic data collection gaps where
missingness is unobservable. Masked Period: a continuous temporal window
of events is removed, simulating extended observational blackouts.
LifeSentence maintains the highest accuracy across all three conditions.
Notably, the Transformer baseline performs worst across all conditions,
falling below even the Logistic regression on Masked History (0.382 vs
0.396), suggesting that the from-scratch Transformer\textquotesingle s
learned representations are more fragile to missing inputs than simpler
models. 95\% confidence intervals (in brackets) were calculated via 1000
bootstrapped samples.
\par}
\end{table}

\begin{figure}[tbp]
{\centering
\includegraphics[width=6.27014in,height=2.77292in]{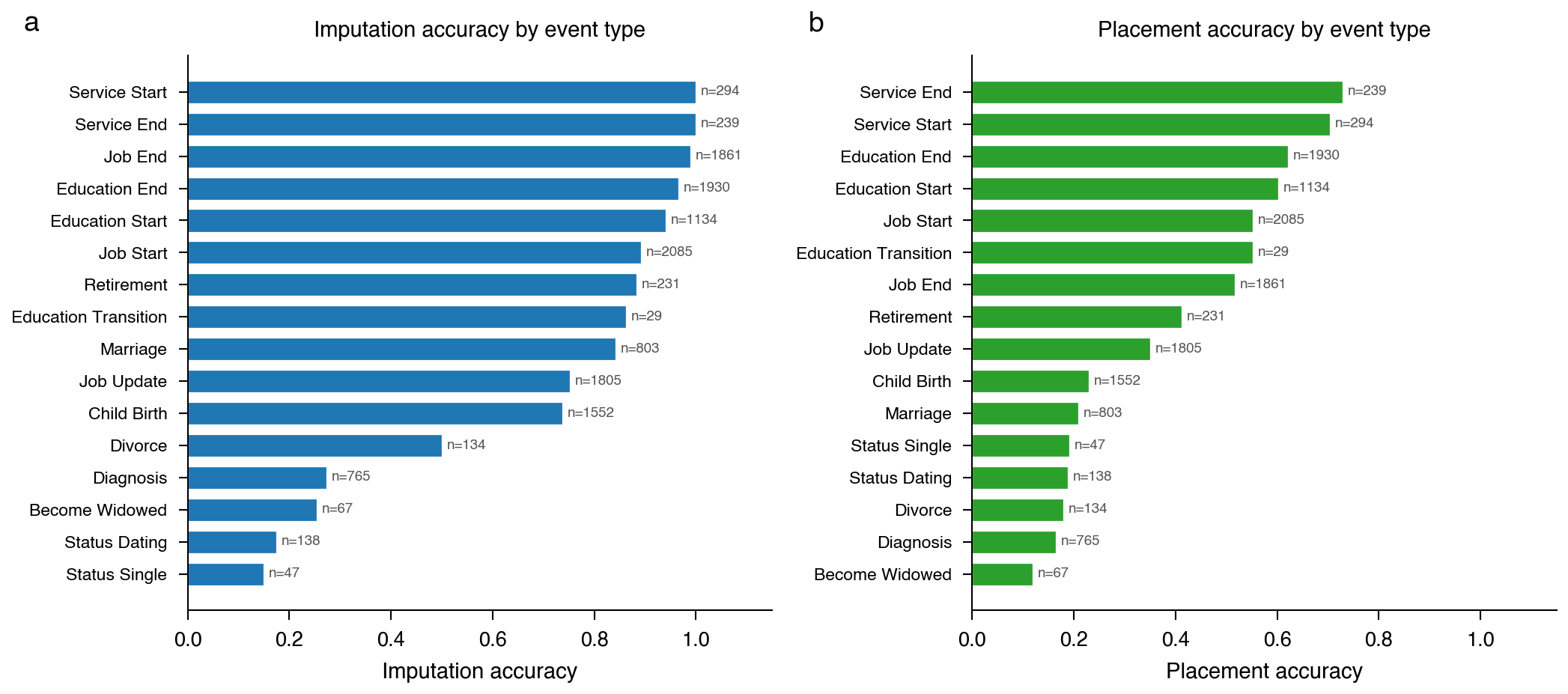}
\par}
\smallskip
{\small
\textbf{Supplementary Figure 10\,---\, Imputation accuracy for
silently deleted events, stratified by event type.} (a) Imputation
accuracy: assessing whether LifeSentence correctly identifies the event
type from the surrounding trajectory context alone. Events embedded in
deterministic institutional sequences --- service end (1.00), service
start (1.00), job end (0.99), education end (0.96) --- are recovered at
near-perfect rates, as their absence creates an unmistakable logical gap
(e.g., a service start without a corresponding service end). Accuracy
declines through events with greater inter-individual variability in
timing and occurrence, such as childbirth (0.74) and divorce (0.50), and
is lowest for high-entropy relational states --- diagnosis (0.27),
widowhood (0.25), dating status (0.17), and single status (0.15) ---
whose presence or absence is weakly constrained by the surrounding event
sequence. (b) Placement accuracy: the proportion of correctly identified
events that are also placed at the correct temporal position within the
trajectory. The rank ordering largely mirrors imputation accuracy ---
institutionally anchored events such as service end (0.73) and service
start (0.70) are easiest to place, while life events with variable
timing such as marriage (0.21), diagnosis (0.16), and widowhood (0.12)
are hardest --- but placement accuracy is uniformly lower, reflecting
the additional difficulty of inferring when a missing event occurred
rather than simply what it was. Together, these panels reveal that
events governed by institutional structure are both identifiable and
temporally localizable from context, while events with higher
inter-individual variability leave weaker contextual traces.
\par}
\end{figure}

\begin{figure}[tbp]
{\centering
\includegraphics[width=6.27014in,height=2.14375in]{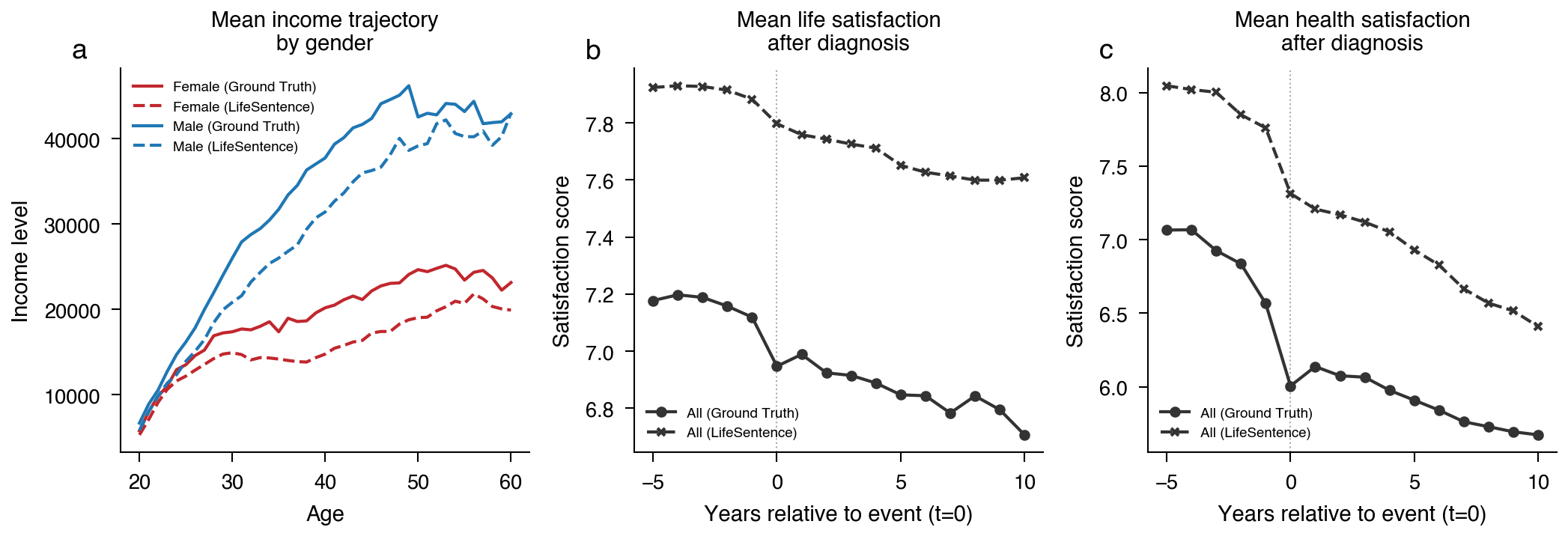}
\par}
\smallskip
{\small
\textbf{Supplementary
Figure 11\,---\, Emergent sociodemographic stratification and
event-driven perturbations in continuous state predictions.} (a) Mean
predicted income trajectories across the lifespan stratified by gender,
comparing ground truth (solid lines) with LifeSentence predictions
(dashed lines) for male (blue) and female (red) individuals. The model
recovers the gender income gap without explicit supervision: male
trajectories exhibit steeper growth through early and mid-career, while
female trajectories follow a flatter curve at lower levels. LifeSentence
preserves the relative magnitude and temporal shape of the divergence
across the working life. (b--c) Mean satisfaction scores in the years
surrounding a diagnosis event (t = 0, dotted vertical line), comparing
ground truth (solid lines with circle markers) with LifeSentence
predictions (dashed lines with cross markers). (b) Life satisfaction
shows a modest decline following diagnosis, which the model captures
directionally while slightly overestimating absolute levels. (c) Health
satisfaction exhibits a sharper and more sustained decline after
diagnosis, with LifeSentence reproducing the overall downward slope and
post-diagnosis trajectory. In both cases, the model systematically
overestimates absolute satisfaction but accurately recovers the
direction, timing, and relative magnitude of diagnosis-induced
perturbations, indicating that the relationship between discrete health
events and subjective well-being is sufficiently structured within
biographical sequences to be learned from observational data alone.
\par}
\end{figure}